\pdfoutput=1

\documentclass[11pt]{article}

\usepackage[]{ACL2023}

\usepackage{times}
\usepackage{latexsym}
\usepackage{amsmath}
\usepackage[T1]{fontenc}

\usepackage[utf8]{inputenc}

\usepackage{microtype}

\usepackage{inconsolata}
\usepackage{microtype}
\usepackage{multirow}
\usepackage{booktabs}
\usepackage{graphicx}
\usepackage{graphics}
\usepackage{makecell}
\usepackage{pifont}
\usepackage{bbold}
\usepackage{color,xcolor,colortbl}
\usepackage{dsfont}
\usepackage{algpseudocode}
\usepackage{pifont}
\usepackage{listings}
\usepackage{hyperref}
\usepackage{breakurl}
\usepackage{multicol}

\definecolor{Gray}{gray}{0.935}
\definecolor{lightgray}{rgb}{0.95, 0.95, 0.95}
\newcolumntype{g}{>{\columncolor{lightgray}}c}

\newcommand*\samethanks[1][\value{footnote}]{\footnotemark[#1]}

\newcommand\mvp{\textsc{MvP}}
\newcommand\unimvp{\textsc{MvP}~{(multi-task)}}

\usepackage[normalem]{ulem}
\useunder{\uline}{\ul}{}

\def\@fnsymbol#1{\ensuremath{\ifcase#1\or *\or \dagger\or \ddagger\or
   \mathsection\or \mathparagraph\or \|\or **\or \dagger\dagger
   \or \ddagger\ddagger \else\@ctrerr\fi}}

\newcommand{\ssymbol}[1]{^{\@fnsymbol{#1}}}

\title{\mvp: Multi-view Prompting Improves Aspect Sentiment Tuple Prediction} %

\author{Zhibin Gou\thanks{~~Equal contribution.}, ~Qingyan Guo\samethanks, ~Yujiu Yang\thanks{~~Corresponding author.}\\
   Tsinghua University \\
  \texttt{zebgou@gmail.com\quad gqy22@mails.tsinghua.edu.cn}\\
  \texttt{yang.yujiu@sz.tsinghua.edu.cn}\\}

\begin{document}
\maketitle

\begin{abstract}

\emph{Generative} methods greatly promote aspect-based sentiment analysis via generating a sequence of sentiment elements in a specified format.
However, existing studies usually predict sentiment elements in a fixed order, which ignores the effect of the interdependence of the elements in a sentiment tuple and the diversity of language expression on the results.
In this work, we propose \emph{\textbf{M}ulti-\textbf{v}iew \textbf{P}rompting} (\mvp) that aggregates sentiment elements generated in different orders, leveraging the intuition of human-like problem-solving processes from different views.
Specifically, \mvp~introduces element order prompts to guide the language model to generate multiple sentiment tuples, each with a different element order, and then selects the most reasonable tuples by voting.
\mvp~can naturally model multi-view and multi-task as permutations and combinations of elements, respectively, outperforming previous task-specific designed methods on multiple ABSA tasks with a single model.
Extensive experiments show that \mvp~significantly advances the state-of-the-art performance on 10 datasets of 4 benchmark tasks, and performs quite effectively in low-resource settings.
Detailed evaluation verified the effectiveness, flexibility, and cross-task transferability of \mvp.\footnote{Code and data released at \url{https://github.com/ZubinGou/multi-view-prompting}}

\end{abstract}

\section{Introduction}

Aspect-based sentiment analysis (ABSA) aims to predict tuples of sentiment elements of interest for a given text. There are four sentiment elements that constitute the main line of ABSA research: aspect term ($a$), aspect category ($c$), opinion term ($o$) and sentiment polarity ($s$) \cite{DBLP:journals/corr/abs-2203-01054}. Given an example sentence, ``I love the sushi badly!'', the corresponding elements are ``sushi'', ``food quality'', ``love'' and ``positive'', respectively.
Early studies focus on a single sentiment element like aspect term \cite{liu-etal-2015-fine, ma-etal-2019-exploring}, aspect category \cite{DBLP:conf/aaai/ZhouWX15} or sentiment polarity \cite{wang-etal-2016-attention, chen-etal-2017-recurrent}.
Recent works propose compound ABSA tasks involving multiple associated elements, such as aspect sentiment triplet extraction (ASTE) \cite{DBLP:conf/aaai/PengXBHLS20}, target aspect sentiment detection (TASD) \cite{DBLP:conf/aaai/WanYDLQP20}, aspect sentiment quad prediction (ASQP) \cite{zhang-etal-2021-aspect} and aspect category opinion sentiment (ACOS) \cite{DBLP:conf/coling/CaiTZYX20}. Their target formats are shown in Table \ref{table:tasks}.

\begin{table}[t]
\centering
\setlength{\tabcolsep}{4pt}
\resizebox{\columnwidth}{!}{%
\begin{tabular}{lrr}
\toprule
Task                                                                                 & Output     \\
\midrule
\begin{tabular}[c]{@{}l@{}}Aspect Category Opinion Sentiment (ACOS)\end{tabular}   & $(a, c, o, s)$   \\
\begin{tabular}[c]{@{}l@{}}Aspect Sentiment Quad Prediction (ASQP)\end{tabular}    & $(a, c, o, s)$ \\
\begin{tabular}[c]{@{}l@{}}Aspect Sentiment Triplet Extraction (ASTE)\end{tabular} & $(a, o, s)$   \\
\begin{tabular}[c]{@{}l@{}}Target Aspect Sentiment Detection (TASD)\end{tabular}   & $(a, c, s)$   \\
\bottomrule
\end{tabular}}
\caption{Aspect sentiment tuple prediction tasks with their corresponding outputs.
Notably, although both ACOS and ASQP are the most complex quadratic prediction tasks, ACOS focuses on implicit aspects and opinions compared to ASQP. Detailed tasks and dataset statistics are shown in Appendix \ref{sec:appendix:data}.}
\label{table:tasks}
\end{table}

\begin{figure*}[t]
  \centering
    \includegraphics[width=1.0\textwidth]{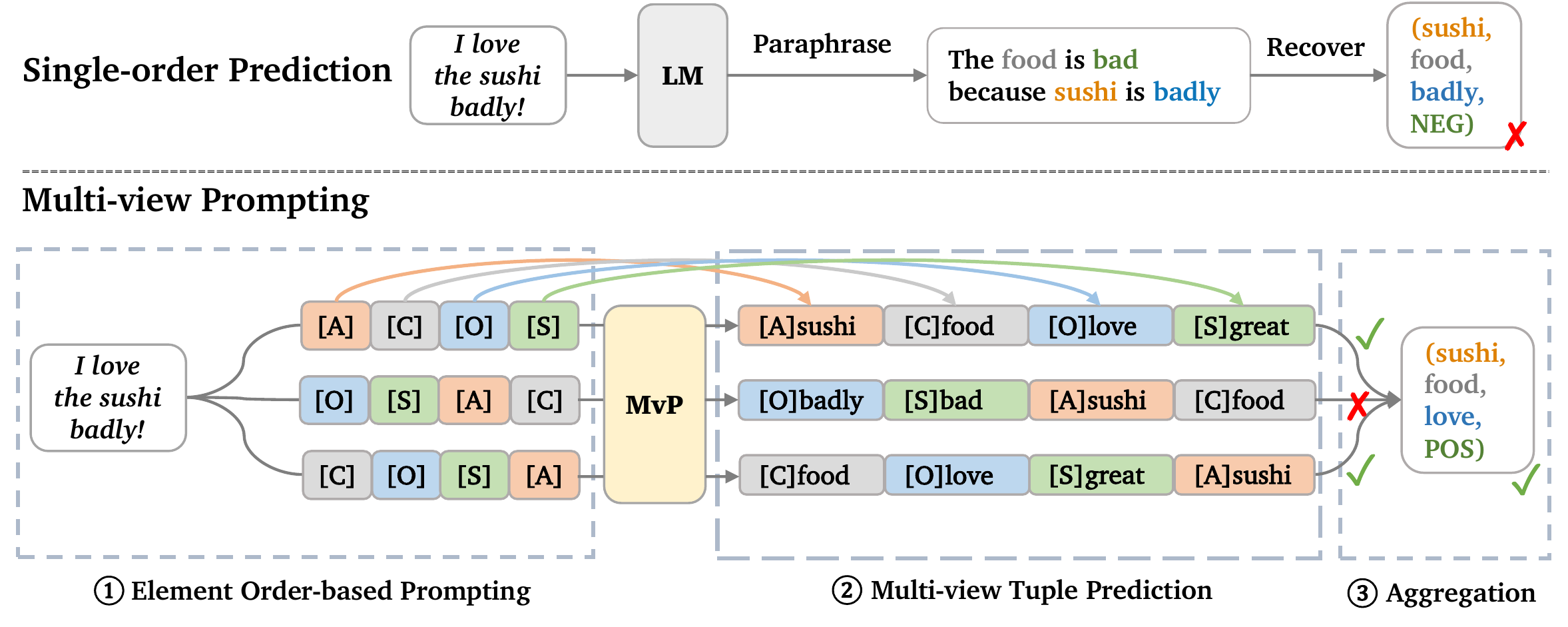}
    \caption{
    Compared with predicting in a single order,
    \mvp~proposes element-order prompt learning to control the prediction order of sentiment element. \mvp~contains three steps: \ding{172} permutes multiple elements to form order prompts and constructs an appropriate subset in terms of conditional generation scores; \ding{173} generates multiple sequences consisting of tuples from different views based on the prompt subset. The element order of each tuple accords with the prompt in the input; \ding{174} aggregates the multiple predictions and obtains the final output.}
    \label{fig:1}
\end{figure*}

Recently, generative methods have been used to handle various ABSA tasks uniformly and achieved good performance \cite{DBLP:journals/corr/abs-2203-01054}, where the common practice is to generate a sequence of sentiment elements in a specified format to leverage label semantics.
To be specific, they use class index \cite{yan-etal-2021-unified}, sentiment element sequence \cite{zhang-etal-2021-towards}, natural language \cite{liu-etal-2021-solving, zhang-etal-2021-aspect-sentiment},  structured extraction schema \cite{lu-etal-2022-unified} or opinion tree \cite{DBLP:conf/ijcai/BaoWJXL22} as the target of the generation models.

However, previous works usually generate the sequence of sentiment elements in a left-to-right fixed order, which ignores the influence of the interdependence of the elements in a sentiment tuple and the diversity of language expression on the targets.
For example, the ``$c\Rightarrow s \Rightarrow a \Rightarrow o$'' order in \textsc{Paraphrase} \cite{zhang-etal-2021-aspect-sentiment} (Figure \ref{fig:1}). 
This single-order generation has the following potential drawbacks:
(1) Incompleteness, tuple prediction is not naturally a text generation task, the relationship among elements is not ordered but interdependent;
(2) Instability, as shown in a study by \citet{hu-etal-2022-improving-aspect}, the performance of different target template orders differs significantly; 
(3) Error accumulation, the previous prediction errors will be accumulated and affect later predictions.

To address the above challenges, we propose \emph{\textbf{M}ulti-\textbf{v}iew \textbf{P}rompting} (\mvp)
that aggregates sentiment elements predicted in different orders,
leveraging the intuition of solving problems from different views in human reasoning and decision \cite{stanovich2000individual}.
Inspired by prompt chaining \cite{DBLP:journals/corr/abs-2107-13586, DBLP:journals/corr/abs-2201-11903, DBLP:journals/corr/abs-2203-11171, DBLP:journals/corr/abs-2207-00747},
\mvp~introduces element order-based prompt learning
to control the prediction order of sentiment elements, enabling diverse target expressions.
Compared to single-order generation, \mvp~mitigates the incompleteness and instability of a fixed order by receiving information from multiple views, while alleviating the potential error accumulation of generative methods via permutation of elements (Figure \ref{fig:1}).
Besides, \mvp~is naturally suited for training a single model to solve multiple ABSA tasks as combinations of elements, adaptively enabling knowledge transfer from related tuple prediction tasks.

We conduct extensive experiments on main aspect sentiment tuple prediction tasks, including ASQP, ACOS, ASTE and TASD.
Empirical results show the superiority of \mvp~in supervised, low-resource, and cross-task transfer settings.
In supervised settings, the single-task and multi-task \mvp~outperform the state-of-the-art by 1.34\% and 1.69\% absolute F1 scores on all tasks, respectively. 
At low resource settings, \mvp~has sizable improvement over strong baselines, and cross-task transfer brings a more remarkable improvement.

Our major contributions are as follows:

    1) We introduce \mvp, an element order-based prompt learning method that improves sentiment tuple prediction by aggregating multi-view results.
    
    2) \mvp~naturally allows us to train a single model simultaneously on all tasks. To the best of our knowledge, the multi-tasking \mvp~is the first single model that substantially outperforms task-specific models on various ABSA tasks.
    
    3) Experiments show that \mvp~significantly advances the state-of-the-art on 10 datasets of 4 tasks and is quite effective in low-resource settings.

\begin{figure*}[t]
  \centering
    \includegraphics[width=1.0\textwidth]{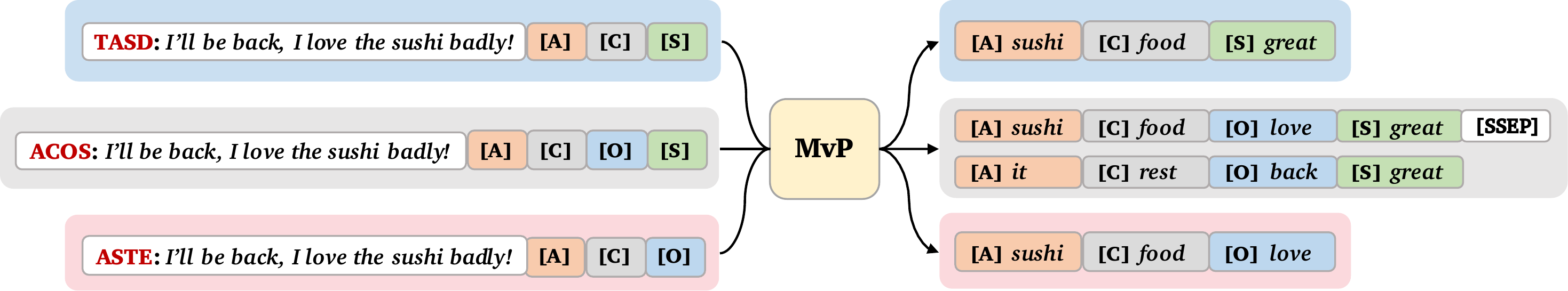}
    \caption{Multi-task learning. \mvp~uniformly tackles ABSA tasks 
    as combination of element order prompts.
    }
    \label{fig:multi-task}
\end{figure*}

\section{Methodology}

To better understand the operation process of the proposed \mvp, we can carefully observe the pipeline shown in Figure \ref{fig:1}. Unlike the fixed order element prediction adopted by previous methods like \textsc{Paraphrase} \cite{zhang-etal-2021-aspect-sentiment}, we take every possible permutation of sentiment elements (6 for the triplet and 24 for the quadruplet) into account and select the appropriate subsets of them for efficiency and effectiveness reasons.
Conditioned on different ordered prompts, a model can generate multiple tuples from different views. Some views give the same correct tuples, while some views are less effective and thus might be wrong, but it's unlikely to result in the same error.
In other words, different views tend to show more agreement in the correct sentiment tuples.
Following this intuition, the proposed \mvp~aggregates and 
takes the tuples that most views agree on as the final result.

\subsection{Problem Definition}

In this section, we present our approach with the quadruple task by default, which can be applied to triplet tasks with minor modifications. We formally define the task as follows:

Given an input sentence, aspect sentiment tuple prediction aims to predict all sentiment tuples $T=\{(a, c, o, 
s)\}$, each consisting of aspect term ($a$), aspect category ($c$), opinion term ($o$) and sentiment polarity ($s$).
To leverage label semantics, following previous works \cite{zhang-etal-2021-aspect}, we paraphrase these elements to natural language ${e_a, e_c, e_o, e_s}$ separately. For example, we map the ``POS'' label of sentiment polarity $s$ to ``great'', and map the ``NULL'' label of opinion term $o$ to ``it''.

\subsection{Element Order-based Prompt Learning}
\label{subsec:method:order-prompt-learning}

To control the prediction order of sentiment elements, \mvp~introduces an element order-based prompting mechanism. Specifically, we design the target with ordered target schema and input with element order prompts.

\subsubsection{Ordered Target Schema}
\label{subsubsec:method:target-schema}
To indicate different sentiment elements, we follow the DLO method \cite{hu-etal-2022-improving-aspect} and design element markers to represent the structure of the information \cite{DBLP:conf/iclr/PaoliniAKMAASXS21}. The element markers for $e_a$, $e_c$, $e_o$, and $e_s$ are $\mathtt{[A]}$, $\mathtt{[C]}$, $\mathtt{[O]}$ and $\mathtt{[S]}$, respectively.
We add the corresponding marker as a prefix to each element and concatenate them in a given permutation $p_i$ as the target sequence,
for example, ``$[\mathtt{O}] e_{o}[\mathtt{A}] e_{a}[\mathtt{C}] e_{c}[\mathtt{S}] e_{s}$''.

If there are multiple sentiment tuples for an input sentence, we utilize a special symbol $\mathtt{[SSEP]}$ to concatenate their corresponding ordered target schema to get the final target sequence $\boldsymbol{y_{p_i}}$.

\subsubsection{Element Order Prompts}
\label{subsubsec:method:order-prompt}
We design element order prompts by concatenating these element markers to represent the desired order $p_i$ of sentiment elements (for example, ``$\mathtt{[O][A][C][S]}$'' indicates prediction in the order of ``$o\Rightarrow a \Rightarrow c \Rightarrow s$''). Then, we add the prompt as a suffix to each input sentence to get the final input $\boldsymbol{x_{p_i}}$. Thus we obtain an input-output pair for training:

\textbf{\emph{Input} ($\boldsymbol{x}$):} I love the sushi badly! $\mathtt{[O][A][C][S]}$

\textbf{\emph{Output} ($\boldsymbol{y}$):} $[\mathtt{O}]$ love $[\mathtt{A}]$ sushi $[\mathtt{C}]$ food $[\mathtt{S}]$ great

We find that the design of element order prompts can effectively guide sentiment tuples' generation order. Thus multi-view and multi-task can be flexibly modeled through the permutation and combination of elements.

\subsection{Multi-view Training}
\label{subsec:method:mult-view-training}

For training, \mvp~selects appropriate element orders to construct input-target pairs, and then fine-tunes a Seq2Seq model.

\subsubsection{Element Order Selection}
\label{subsubsec:method:order-select}
Since overheads increase linearly with the number of views and the performance of different views varies, we need to select appropriate element orders.
Following the study of prompt ordering \cite{lu-etal-2022-fantastically, hu-etal-2022-improving-aspect}, we choose the potentially better-performing orders based on the average entropy of the candidate permutations on the training set. The steps are as follows:
(i) we use every possible permutation $p_i$ of sentiment elements as candidates;
(ii) given an input sentence $\boldsymbol{x}$ and its target tuples, we construct the ordered target schema $\boldsymbol{y_{p_i}}$ of permutation $p_i$ as described in \S \ref{subsubsec:method:target-schema}, replace the element markers in it with spaces to avoid noises, and query a pre-trained language model to get conditional generation scores $p(\boldsymbol{y_{p_i}}|\boldsymbol{x})$;
and (iii) calculate the average score of permutation $p_i$ over the training set $D$: 
\begin{equation}
     S_{p_i}=\frac{\sum_{D}p(\boldsymbol{y_{p_i}}|\boldsymbol{x})}{|D|}   
\end{equation}

Thus we can rank each permutation $p_i$ with $S_{p_i}$ and top \emph{m} permutations are used for training.

\subsubsection{Training}
\label{subsubsec:method:training}
With the selected $m$ permutations, we construct $m$ different ordered prompts and targets for each sentence. Given the input-target pair $(\boldsymbol{x}, \boldsymbol{y})$, we can fine-tune a pre-trained sequence-to-sequence language model (LM) such as BART \cite{lewis-etal-2020-bart} or T5 \cite{DBLP:journals/jmlr/RaffelSRLNMZLL20}, minimizing the following negative log-likelihood loss:
\begin{equation}\label{nll-loss}
\begin{aligned}
\mathcal{L}_{NLL} &= -\mathbb{E} \log p(\boldsymbol{y} | \boldsymbol{x}) \\ 
 &= -\mathbb{E}\sum_{t=1}^{T}\log p(\boldsymbol{y_t} | \boldsymbol{x}, \boldsymbol{y_{<t}})
\end{aligned}
\end{equation}
where $T$ is the length of the target sequence $\boldsymbol{y}$ and $\boldsymbol{y_{<t}}$ denotes previously generated tokens.

\subsection{Multi-view Inference}
\label{subsec:method:mvp}

For inference, \mvp~prompts the trained model to normatively generate multiple sentiment tuples in previously selected orders, and finally aggregate to obtain the most reasonable tuples. 

\subsubsection{Schema Constrained Generation} 
\label{subsubsec:method:constrained-decoding}
Given an input sentence, we construct multiple prompts in the same order as the training, which guides the model to generate targets from different views. 
However, the generated results may not conform to the target schema format, especially when the training set is small \cite{zhang-etal-2021-aspect, yan-etal-2021-unified}. Therefore, we designed a schema-based constrained decoding \cite{DBLP:conf/iclr/CaoI0P21} that injects target schema knowledge into the decoding process. It ensures that the generated elements are in the corresponding vocabulary set. See Appendix \ref{sec:appendix:cd} for implementation details.

\subsubsection{Multi-view Results Aggregation}
\label{subsubsec:method:aggregation}
Since each view may predict more than one tuple, we first aggregate the results of all views and then use the tuples that appear in most views as the final prediction.
Specifically, for an input sentence $\boldsymbol{x}$, suppose we prompt a trained model to generate from $m$ selected permutations, and the set of predicted tuples for permutation $p_i$ is $T'_{p_i}$, which may contain one or more sentiment tuples, and then we can obtain the final aggregated result $T_{\textit{\mvp}}'$ by the following equation:
\begin{equation*}
    T_{\textit{\mvp}}'=\{t| t\in \bigcup_{i=1}^{m} T'_{p_i} ~\text{and}~(\sum_{i=1}^m{\mathds{1}_{T'_{p_i}}(t)} \geq \frac{m}{2})\}
\end{equation*}

\begin{table*}[t]
\centering
\setlength{\tabcolsep}{4pt}
\resizebox{\textwidth}{!}{%
\begin{tabular}{l||cc|cc|cc|cccc||g}
\toprule
\multirow{2}{*}{\textbf{Methods}} & \multicolumn{2}{c|}{\textbf{ASQP}} & \multicolumn{2}{c|}{\textbf{ACOS}} & \multicolumn{2}{c|}{\textbf{TASD}} &  \multicolumn{4}{c||}{\textbf{ASTE}} & \cellcolor{lightgray}\ \\
               & \textbf{R15} & \textbf{R16} & \textbf{Lap} & \textbf{Rest} & \textbf{R15} & \textbf{R16} & \textbf{L14} & \textbf{R14} & \textbf{R15} & \textbf{R16} & \multirow{-2}{*}{\textbf{AVG}} \\
\midrule
TAS-BERT \small{\cite{DBLP:conf/aaai/WanYDLQP20}} & 34.78  & 43.71  & 27.31  & 33.53 & 57.51  & 65.89 & - & - & - & - & - \\
Jet-BERT \small{\cite{xu-etal-2020-position}}  & - & - & - & - & - & - & 51.04 & 62.40 & 57.53 & 63.83 & - \\
Extract-Classify \small{\cite{cai-etal-2021-aspect}} &  36.42  & 43.77  & 35.80 & 44.61 & - & - & - & - & - & - & - \\
GAS \small{\cite{zhang-etal-2021-towards-generative}} & 45.98  & 56.04  & - & - & 60.63  & 68.31 & 58.19 & 70.52 & 60.23 & 69.05 & - \\
Paraphrase \small{\cite{zhang-etal-2021-aspect-sentiment}} & 46.93  & 57.93  &  {43.51}  & \underline{61.16} & 63.06  & \underline{71.97} & 61.13 & 72.03 & 62.56 & 71.70 & 61.20 \\
UIE \small{\cite{lu-etal-2022-unified}} & - & - & - & - & - & - & 62.94  & 72.55 & 64.41 & 72.86 & - \\ 
Seq2Path \small{\cite{mao-etal-2022-seq2path}} &   -    &  -  &   42.97  & 58.41 & 63.89 & 69.23 & \underline{64.82} & \underline{75.52} & \underline{65.88} & 72.87 & - \\  %
DLO \small{\cite{hu-etal-2022-improving-aspect}} & 48.18  & \underline{59.79}  &  43.64 &  59.99 & 62.95 &  71.79 & 61.46 & 72.39 & 64.26 & {73.03} & 61.75 \\
\midrule
UnifiedABSA$\ssymbol{2}$\small{\cite{DBLP:journals/corr/abs-2211-10986}} &    -     &  -  &   42.58  & {60.60} & - & - & - & - & - & - & - \\ 
LEGO-ABSA$\ssymbol{2}$\small{\cite{gao-etal-2022-lego}} &  46.10  &  57.60  &   -  &   - & 62.30  &  71.80 & 62.20 & 73.70 & 64.40 &  69.90 & - \\ 
\midrule
\mvp & \underline{51.04} & \textbf{60.39}  &  \textbf{43.92} & \textbf{61.54} & \underline{64.53} & \textbf{72.76} & 63.33  & {74.05} & \underline{65.89} & \textbf{73.48} & \underline{63.09}\\
\unimvp $\ssymbol{2}$ & \textbf{52.21} & 58.94  & \underline{43.84}  & 60.36 & \textbf{64.74} & 70.18 & \textbf{65.30} & \textbf{76.30}  & \textbf{69.44} & \underline{73.10} & \textbf{63.44}\\
\bottomrule
\end{tabular}
}
\caption{\label{asqp-result}
Main results on 10 datasets of ASQP, ACOS, TASD and ASTE tasks. F1 scores are reported; the best results are in bold, while the second best are underlined. 
$\ssymbol{2}$ indicates multi-tasking models. 
}
\end{table*}

\begin{table*}[t]
\centering
\setlength{\tabcolsep}{4pt}
\resizebox{1\textwidth}{!}{%
\begin{tabular}{c||l|c||ccccc||g}
\toprule
 & \textbf{Methods} & \textbf{Transfer Source}  & \textbf{1\%} & \textbf{2\%} & \textbf{5\%} & \textbf{10\%} & \textbf{20\%} & \textbf{AVG} \\ \midrule
\multirow{5}{*}{\textbf{\begin{tabular}[c]{@{}c@{}}ASQP\\ (R15)\end{tabular}}}
& Paraphrase \small{\cite{zhang-etal-2021-aspect-sentiment}}& -  & 5.90 & 15.73 & 24.16 & 31.33 & 37.47 & 22.92 \\
& DLO \small{\cite{hu-etal-2022-improving-aspect}} & -  & 10.03 & 15.94 & 29.13 & 35.89 & 40.34 & 26.27 \\
& \mvp & -& {\bf 13.46} & {\bf 22.58} & {\bf 32.44} & {\bf 38.48} & {\bf 41.82} & {\bf 29.76} \\ \cmidrule{2-8}
& DLO (transfer) & ASTE (R15) & 26.28 & 28.72 & 35.94 & 39.48 & 42.92 & 34.67   \\
& \mvp~{(transfer)} & ASTE (R15) & \textbf{28.69} & \textbf{33.93} & \textbf{40.08} & \textbf{43.10} & \textbf{45.09} & \textbf{38.18}  \\ \midrule \midrule
 
\multirow{5}{*}{\textbf{\begin{tabular}[c]{@{}c@{}}ACOS\\ (Rest)\end{tabular}}}
 & Paraphrase \small{\cite{zhang-etal-2021-aspect-sentiment}} & - & 14.85 & 24.81 & 38.33 & 45.32 & 49.64 & 34.59 \\
 & DLO \small{\cite{hu-etal-2022-improving-aspect}} & - & 19.84 & 29.84 & 38.47 & 43.45 & 46.47 & 35.61 \\
 & \mvp  & - & {\bf 23.84} & {\bf 32.57} & {\bf 42.89} & {\bf 47.77} & {\bf 53.54} & {\bf 40.12} \\ \cmidrule{2-8}
 & DLO (transfer) & ASTE (R16)& 31.06 & 40.55 & 43.23 & 45.74 & 47.98 & 41.71 \\
 & \mvp~(transfer) & ASTE (R16)& \textbf{39.24} & \textbf{42.72} & \textbf{49.78} & \textbf{52.53} & \textbf{55.28} & \textbf{47.91}  \\ \midrule \midrule
 
\multirow{5}{*}{\textbf{\begin{tabular}[c]{@{}c@{}}TASD\\ (R16)\end{tabular}}}
 & Paraphrase \small{\cite{zhang-etal-2021-aspect-sentiment}} & - & 26.29 & 36.70 & 49.48 & 55.66 & 61.79 & 45.98 \\
 & DLO \small{\cite{hu-etal-2022-improving-aspect}}& -  & 29.66 & 41.17 & 50.44 & 58.27 & 62.43 & 48.39 \\
 & \mvp & - & {\bf 34.00} & {\bf 41.76} & {\bf 52.58} & {\bf 58.93} & {\bf 64.53} & {\bf 50.36} \\ \cmidrule{2-8}
 & DLO (transfer) & ASQP (R16) & 66.25 & 66.21 & 64.54 & 67.99 & 68.50 & 66.70 \\ 
 & \mvp~{(transfer)} & ASQP (R16) & \textbf{68.49} & \textbf{68.06} & \textbf{68.47} & \textbf{68.98} & \textbf{69.89} & \textbf{68.78} \\ \midrule \midrule
 
\multirow{5}{*}{\textbf{\begin{tabular}[c]{@{}c@{}}ASTE\\ (L14)\end{tabular}}}
 & Paraphrase \small{\cite{zhang-etal-2021-aspect-sentiment}}& -  & 16.29 & 29.20 & 38.61 & 45.20 & 52.88 & 36.44 \\
 & DLO \small{\cite{hu-etal-2022-improving-aspect}} & - & 17.07 & 26.07 & 38.92 & 48.85 & 53.82 & 36.95 \\
 & \mvp & - & {\bf 28.17} & {\bf 34.38} & {\bf 42.89} & {\bf 52.33} & {\bf 54.60} & {\bf 42.47} \\ \cmidrule{2-8}
 & DLO (transfer)$\ssymbol{3}$ & ASQP (R16) & 44.76 & 48.86 & 51.22 & \textbf{56.43} & 56.71 & 51.60 \\
 & \mvp~{(transfer)}$\ssymbol{3}$ & ASQP (R16) & \textbf{48.43} & \textbf{50.33} & \textbf{54.27} & {56.34} & \textbf{59.05} & \textbf{53.68} \\\bottomrule
\end{tabular}
}
\caption{\label{tab:few}
Low-resource and cross-task transfer results. We cover 4 tasks, 4 datasets and 2 domains. For a fair comparison, here we choose DLO-top5 which augments the original training set by 5 times. 
In cross-task transfer settings,
for quadruplet tasks ASQP and ACOS, we first train the model on ASTE (R15) and ASTE (R16), respectively, while for triplet tasks TASD and ASTE, the transfer source is ASQP (R16).
Then we vary the percentage of transfer target training set and report the results.
$\ssymbol{3}$
It is notable that transferring setting on ASTE (L14) is cross-domain, from Restaurant to Laptop. 
}
\end{table*}

\section{Experiments}
\label{sec:exps}

\subsection{Tasks and Dataset}
We validate our methods on 10 datasets over 4 tasks, including quadruplet tasks, ASQP and ACOS, and triplet tasks, ASTE and TASD. For a fair comparison, we apply the same data splits as previous works. The targets of each task are shown in Table \ref{table:tasks} and the detailed statistics are in Appendix \ref{sec:appendix:data}. 

For the ASQP task, we adopt two datasets in the restaurant domain based on SemEval tasks \cite{pontiki-etal-2015-semeval, pontiki-etal-2016-semeval}, Rest15 and Rest16 aligned and completed by \citet{zhang-etal-2021-aspect} subsequently. For the ACOS task, we apply Restaurant-ACOS and Laptop-ACOS constructed by \citet{cai-etal-2021-aspect}. Compared with ASQP, datasets of ACOS focus on implicit aspects and opinions, which helps to measure our methods comprehensively.
For the triple tasks, we adopt datasets provided by \citet{xu-etal-2020-position} and \citet{DBLP:conf/aaai/WanYDLQP20} for ASTE \cite{DBLP:conf/aaai/PengXBHLS20} and TASD, respectively.

\subsection{Implement Details}
We employ T5-\textsc{Base} model \cite{DBLP:journals/jmlr/RaffelSRLNMZLL20} from Huggingface Transformers library\footnote{\url{https://github.com/huggingface/transformers}}\cite{wolf-etal-2020-transformers} as the pre-trained model. T5 adopts a classical encoder-decoder architecture similar to Transformer \cite{DBLP:conf/nips/VaswaniSPUJGKP17}.
We use greedy search for decoding by default. We use the same hyperparameters across all tasks and datasets, and detailed settings can be found in Appendix \ref{sec:appendix:exps}.

The number of views $m$ is set to 5 by default across the majority of the experiments, including multi-task, low-resource, cross-task transfer, and ablations.
Only the single-task model in the main experiment uses 15 views for the quadruplet tasks and 5 views for the triplet tasks. For simplicity, the number of views in inference is the same as that in training. The case of using a different number of views is left for further exploration.

In the multi-task settings, 
to introduce domain information,
we simply add the task name and dataset name followed by colon separators (e.g. ``ASQP: Rest15: '') as the prefixes to each input sentence, and train a single model on all datasets across all tasks (ASQP, ACOS, TASD, ASTE).
We select the appropriate element orders for each dataset separately.
We find overlap between the input sentences across different splits of different datasets. Therefore, to avoid data leakage, we collect the training sets from all datasets, and discard samples that overlap with the test set of any task. Then we split the data by 9:1 to obtain the final training and validation sets for our multi-tasking method.

\subsection{Evaluation Metrics}

For all ABSA tasks, a predicted sentiment tuple is considered as correct if and only if all its elements are exactly the same as the gold tuple. We use F1 scores as the main evaluation metrics \cite{zhang-etal-2021-aspect, mao-etal-2022-seq2path}.
All reported F1 scores are averaged over 5 runs with different random seeds. 
For multi-task settings, we use a different split of the training and development sets in each run.

\subsection{Compared Methods}
\label{sec:compared-methods}
We compare our methods with the following three types of previous state-of-the-art methods:

    \textbf{Discriminative methods.}
    \textbf{TAS-BERT}, based on extraction, \cite{DBLP:conf/aaai/WanYDLQP20} jointly detects the sentiment tuples. \textbf{Extract-Classify} \cite{cai-etal-2021-aspect} decomposes the ACOS task into two steps. For ASTE, \textbf{Jet-BERT} \cite{xu-etal-2020-position} addresses the task in an end-to-end framework by a tagging scheme.
    
    \textbf{Generative methods.}
    \textbf{GAS} \cite{zhang-etal-2021-towards-generative} is the first to model ABSA tasks as a generation process. \textbf{Paraphrase} \cite{zhang-etal-2021-aspect} designs semantic templates filled with fixed-order elements of tuples as generation targets. \textbf{Seq2Path} \cite{mao-etal-2022-seq2path} generates tuples as paths of a tree and then selects valid ones. \textbf{DLO / ILO} \cite{hu-etal-2022-improving-aspect}
    augments ASQP dataset given the order-free property of the quadruplet based on templates. 
    We also consider \textbf{UIE} \cite{lu-etal-2022-unified}, a unified text-to-structure framework to model various IE tasks which is pre-trained on large-scale data.

    \textbf{Multi-tasking methods.}
    A recent trend is tackling multiple ABSA tasks uniformly using a single multi-tasking model. \textbf{LEGO-ABSA} \cite{gao-etal-2022-lego} designs task prompts similar to T5 and \textbf{UnifiedABSA} \cite{DBLP:journals/corr/abs-2211-10986} adopts instruction tuning \cite{mishra-etal-2022-cross, DBLP:conf/iclr/WeiBZGYLDDL22}.

As a fair comparison, all results of these supervised methods are obtained from the base pre-trained model, either BERT or T5.

    \textbf{Large language model (LLM).}
    To compare our method with advanced large language models, we additionally include evaluation results of \textbf{ChatGPT}\footnote{\url{https://chat.openai.com/}} (\texttt{gpt-3.5-turbo})  on the four ABSA tasks with zero- and few-shot prompts. 
    The results can be found in Appendix \ref{sec:appendix:chatgpt}.

\section{Results and Discussions}

\subsection{Single-task and Multi-task Results}
\label{subsec:results:supervised}
Our methods outperform previous best baselines significantly in supervised settings among 4 tasks, 10 datasets, becoming the new state-of-the-art in all of them. As shown in Table \ref{asqp-result}, we observe that:

 1) \textit{By aggregating results from multiple views, \mvp~surpasses the most of previous single-order methods.}
 In comparison with Paraphrase, the single-view method which applies templates with elements in a fixed order, \mvp~achieves a sizable improvement of 1.89\% on average, verifying the effectiveness of multiple informative views. 
 
2) \textit{Element order-based prompt learning effectively guides the generation of tuples by unifying training and inference.} Compared with DLO augmenting data on the target side, \mvp~obtains an improvement of 1.34\% by generating tuples controllably with designed element order-based prompts. 

3) \textit{\mvp~can be applied without abundant pre-training simply and achieves better performance.}
It is notable that \mvp~trained on T5-base exceeds UIE using T5-v1.1-base and subsequently trained on a large corpus with 65M instances on all datasets of ASTE (+1.00 on average).

\textbf{Multi-task Learning.} 
\textit{By permutation and combination, \unimvp~obtains generalized ability among diverse tasks.}
Compared with LEGO-ABSA, a multi-task unified baseline, \unimvp~obtains a +2.82\%  absolute improvement in F1 score on average. 

\mvp, a single model completing tuple prediction by
information from multiple views, can be naturally employed on multiple ABSA tasks, achieving competitive or better performances than previous methods that require task-specific fine-tuning, data augmentation (e.g., Seq2Path) or complex pre-training (e.g., UIE), either in single-task or multi-tasking settings, showing strong stability.

\subsection{Low-resource Results}
\label{subsec:results:low-resource}
To further explore the behavior of our methods in low-resource settings, we train Paraphrase, DLO, \mvp~using 1\%, 2\%, 5\%, 10\%, and 20\% of 4 different training sets in 2 domains over 4 tasks. The F1 scores of test sets are reported in Table \ref{tab:few}. 
We find that \mvp, with efficient prompts from different views, achieves better results than previous works with only a small number of samples. In particular, \mvp~outperforms Paraphrase and DLO substantially in all settings with a performance boost of 5.70\% and 3.87\% F1 on average.

\textbf{Cross-task transfer.}
\textit{Based on \mvp, transfer brings further significant improvements, from triplets to quadruplets and vice versa.} \mvp~(transfer) performs relatively well in extremely low-resource situations, thus exceeding strong baselines under cross-task transfer situations, both in-domain and cross-domain. 
Compared with DLO (transfer), \mvp~(transfer) achieves considerably better results under various transfer settings, showing a strong transferability (from 50.36\% to 68.78\% for TASD). 
Rather than capture task-specific features, \mvp~effectively shares ABSA abilities. 
\mvp~(transfer) trained with simple tasks (TASD and ASTE here) with adequate data can be easily transferred to tough tasks (ASQP and ACOS here) when the dataset sizes are small, and vice versa.

\begin{figure}[t]
  \centering
    \includegraphics[width=\columnwidth]{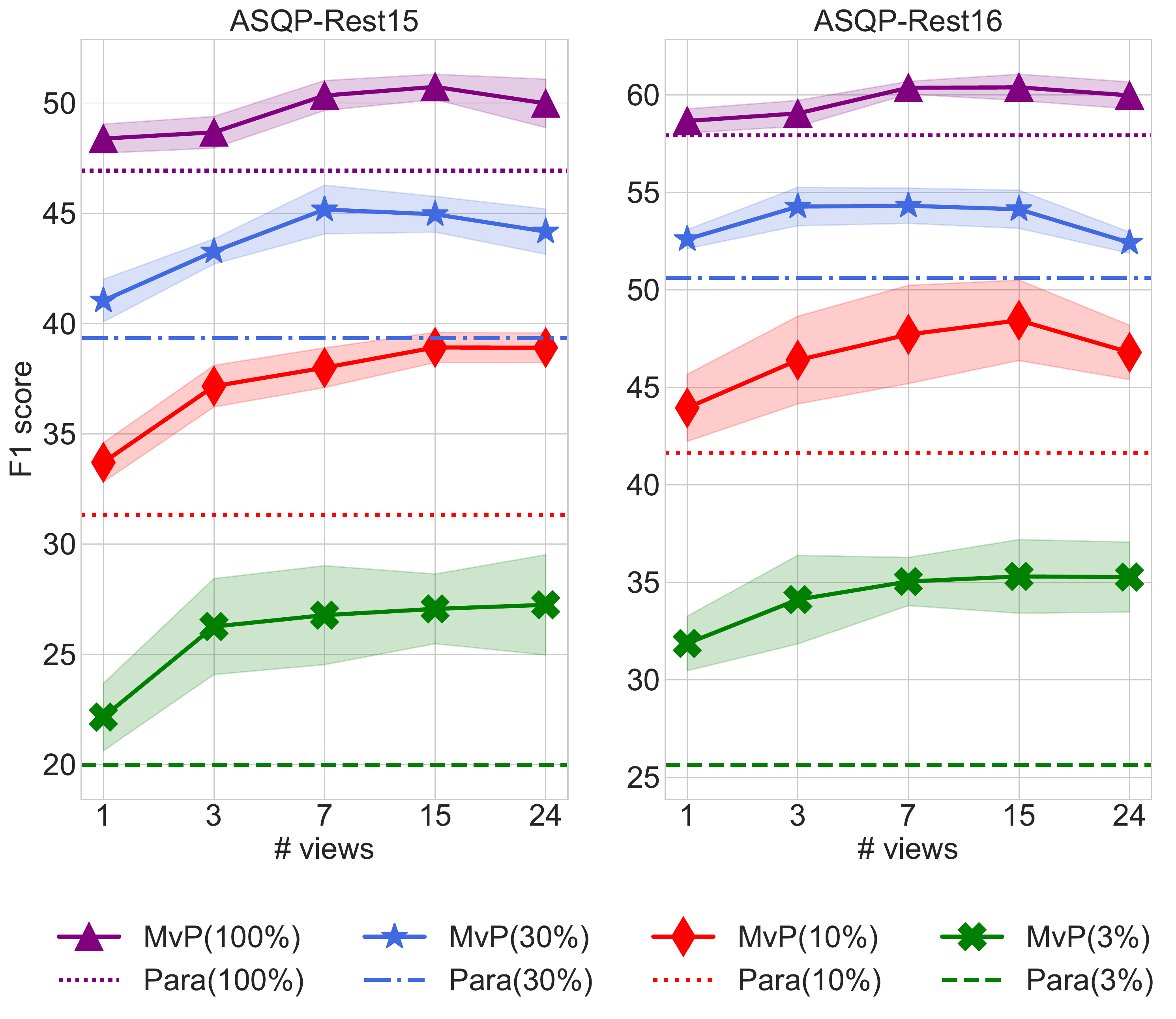}
    \caption{Effect of the number of views. ``\# views'' refers to the number of views used for training and inference, and ``Para'' stands for Paraphrase.
    Values in parentheses represent the ratio of the training data used. }
    \label{fig:num-view}
\end{figure}
\begin{table}[t]
  \centering
  \setlength{\tabcolsep}{4pt}
\resizebox{\columnwidth}{!}{%
  {
    \begin{tabular}{l||ccc|ccc}
      \toprule
      \multirow{2}{*}{\textbf{Methods}} & \multicolumn{3}{c|}{\textbf{ASTE (L14)}} & \multicolumn{3}{c}{\textbf{ASQP (R15)}}  \\
                 & 1\% & 10\% & 100\% & 1\% & 10\% & 100\% \\
      \midrule
      \mvp~w/o cd  & 21.37 & 49.98 & \underline{63.27} & 12.09  & \underline{37.87} &  \underline{50.92}  \\ \midrule
      \mvp~(rand)  & \underline{27.32} & \underline{51.02} & 62.50 & \textbf{13.56}  & 37.18 &  49.84   \\
      \mvp~(rank)  & 25.98 & 49.98 & 62.48 & 13.38  & 37.45 &  49.98      \\
      \midrule
      \mvp       & \textbf{28.37} & \textbf{52.33} & \textbf{63.33} & \underline{13.46} & \textbf{38.48}  & \textbf{51.04}  \\
      \bottomrule
    \end{tabular}
  }
  }
  \caption{
    Ablation of constrained decoding and effect of aggregation strategy on ASTE (L14) and ASQP (R15). “w/o cd” discards the constrained decoding during inference. \mvp~(rank) and \mvp (rand) are both single-view strategies. The former selects the top-ranked sequence based on the prediction scores (perplexity) of generated sequences from multiple views during inference while the latter randomly samples one. }
  \label{tab:ablation}%
\end{table}

\begin{figure*}[t]
  \centering
    \includegraphics[width=\textwidth]{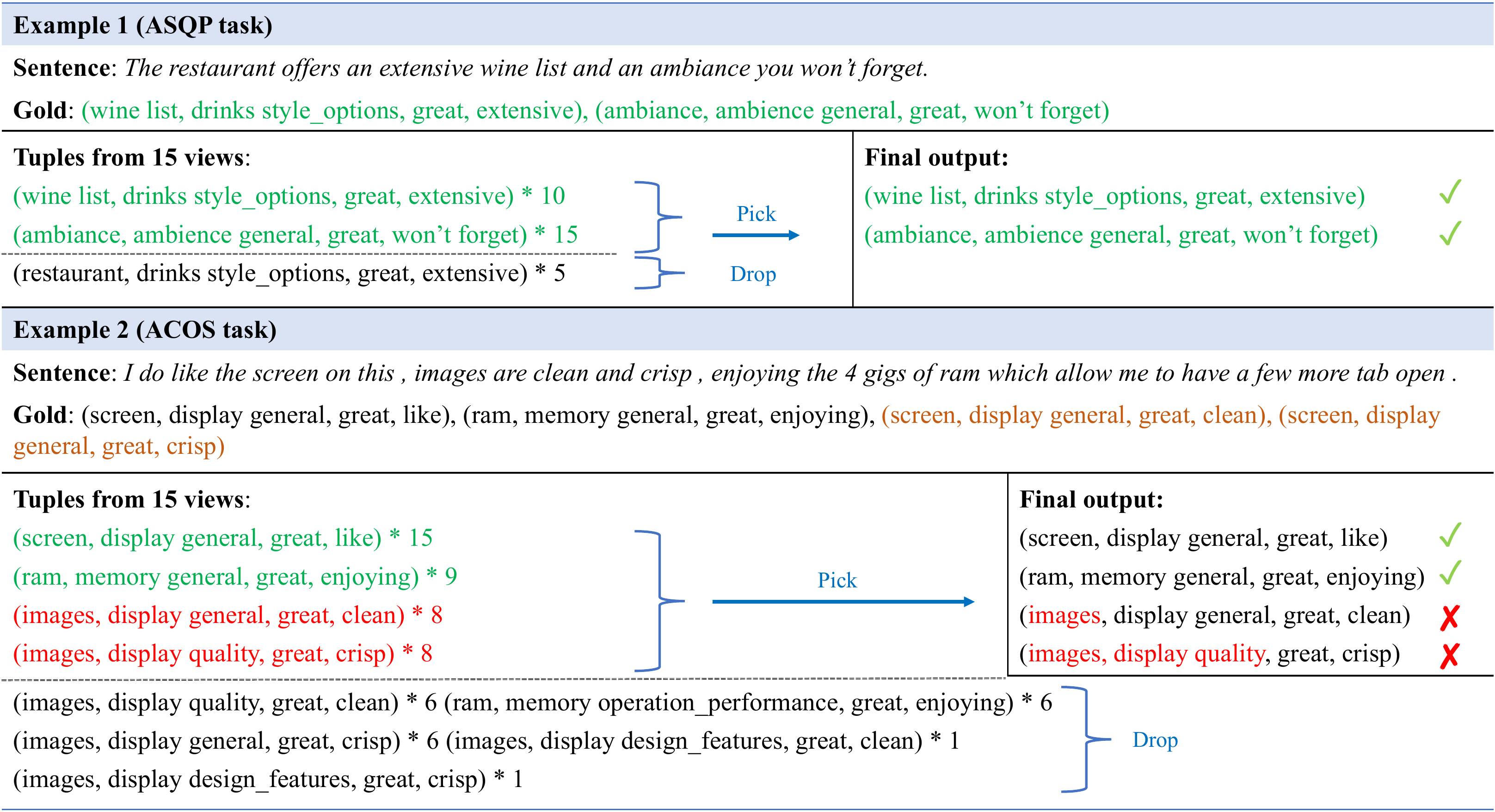}
    \caption{Two examples including the input sentence, quadruplets or triples predicted, and the final outputs of \mvp~after filtering by voting. \textit{Pick} means that the tuple has appeared in more than half of the predictions in multiple views, while \textit{drop} means that it has appeared less than half of the times and is discarded. Words in \textcolor[RGB]{78,173,91}{green} are positive ones while those in \textcolor[RGB]{255,0,0}{red} are wrongly picked. Tuples in \textcolor[RGB]{197,90,17}{orange} are the ones that \mvp~ignores.
    }
    \label{fig:case}
\end{figure*}

\subsection{Effect Analysis}
\label{sec:results:ablation}

\textbf{Effect of the number of views.}
\mvp~raises a question that how many views should be selected for training and inference, which we further explore by varying the number of views and the size of the training set  (Figure \ref{fig:num-view}). 
As the number of views increases, curves show an ascending trend first. Interestingly, when the resource is adequate, F1 decreases slightly after a certain number (between 7 and 15). We believe views with lower ranks may be less effective. Thus it is crucial to balance the size of the data and the number of views. 
It is dramatic that \mvp~performs decently even with a single view, probably due to the appropriate order selection and the constrained decoding. In extremely low-resource scenarios, 
by setting a larger value, \mvp~can expand single-view information and provide more potential choices. The maximum number of views for quadruplets is much higher than that for triples, making \mvp~more appropriate for quadruple tasks.
We provide further comparisons with single-view prompting (i.e., selecting the best single view for training and inference) on all tasks in the Appendix \ref{sec:appendix:svp}.

\textbf{Effect of aggregation strategy.} 
To explore the effect of different aggregation strategies, 
we conduct ablation studies mainly on ASTE and ASQP tasks, as shown in Table \ref{tab:ablation}.
We can see that replacing majority voting with random selection or ranking results in a reduction of F1 in most cases, indicating that majority voting is a more stable strategy for handling diverse views.

\textbf{Effect of constrained decoding.}
The designed constrained decoding guides the generation of different views
by limiting the predicted term to a specific list.
The impact of this algorithm increases as the size of the data decreases, and in extremely low-resource scenarios, \mvp~combined with this algorithm performs considerably well (Table \ref{tab:ablation}).

\subsection{Case Study \& Error analysis}
Figure \ref{fig:case} shows two examples in Rest16 ASQP and Laptop-ACOS, respectively. It can be observed from Example 1 that  \mvp~handles cases with multiple sentiment tuples in a sentence well after filtering unreasonable tuples predicted, i.e. (\textit{restaurant, drinks style\_options, great, extensive}), appearing in five generated results in the case. 
\mvp~only outputs tuples considered important in most views and thus repairs the error in the single view by receiving and aggregating information from multiple views. In Example 2, the challenging Laptop dataset includes 121 categories, and we can see that while multi-view prompting provides more possibilities and choices, it still confuses similar aspect categories, i.e., \textit{display general} and \textit{display quality}.

\section{Related Works}

\textbf{Aspect-base Sentiment Analysis.}
ABSA has received wide attention in recent years.
Early studies focused on extracting or predicting a single sentiment element like aspect term extraction \cite{qiu-etal-2011-opinion, liu-etal-2015-fine, ma-etal-2019-exploring}, aspect category detection \cite{DBLP:conf/aaai/ZhouWX15, bu-etal-2021-asap} or sentiment polarity classification for a given aspect \cite{wang-etal-2016-attention, chen-etal-2017-recurrent, lei-etal-2018-multi, DBLP:conf/aaai/LeiYYZG019}.
Some works further consider the joint prediction of two associated elements \cite{cai-etal-2020-aspect}, including aspect-opinion pair extraction \cite{DBLP:conf/aaai/WangPDX17, chen-etal-2020-synchronous}, aspect term-polarity co-extraction \cite{huang-carley-2018-parameterized, luo-etal-2019-doer, chen-qian-2020-relation}. 
And recent works propose more challenging ABSA tasks to predict sentiment triplets or quadruplets \cite{chen-etal-2022-enhanced}, the most influential of which are ASTE \cite{DBLP:conf/aaai/PengXBHLS20, zhai-etal-2022-com}, TASD \cite{DBLP:conf/aaai/WanYDLQP20}, ASQP \cite{zhang-etal-2021-aspect} and ACOS with an emphasis on the implicit aspects or opinions \cite{DBLP:conf/coling/CaiTZYX20}.

\textbf{Generative ABSA.}
Instead of separate or pipeline methods \cite{phan-ogunbona-2020-modelling}, most recent works attempt to tackle various ABSA problems using a unified framework \cite{DBLP:journals/ijautcomp/SunLQH22}. Generative methods achieve good performance in ABSA by mitigating the potential error propagation in pipeline methods and fully exploiting the rich label semantic information \cite{DBLP:conf/iclr/PaoliniAKMAASXS21, DBLP:journals/corr/abs-2203-01054, yu2023syngen}.
They use sentiment element sequence \cite{zhang-etal-2021-towards}, natural language \cite{liu-etal-2021-solving, zhang-etal-2021-aspect-sentiment} and  structured extraction schema \cite{lu-etal-2022-unified} etc. as the generative targets.
Recently proposed LEGO-ABSA \cite{gao-etal-2022-lego} and UnifiedABSA \cite{DBLP:journals/corr/abs-2211-10986} focus on multi-tasking with task prompts or instruction design.
\citet{hu-etal-2022-improving-aspect} firstly investigate element ordering and propose methods to augment target-side data with selected orders for the ASQP task.
Despite the promising results, the augmentation may confuse the model with multiple targets for the same input (i.e., one-to-many), thus leading to discrepancies between inference and training.
We fill the gap and eliminate such confusion by aligning training and inference with multi-view prompt learning.

\section{Conclusion}

In this work, we introduce an element order-based prompt learning method - \mvp, which improves aspect-level opinion information prediction by simple yet effective multi-view results aggregation.
Leveraging the intuition of solving problems from different views, \mvp~advances the research of generative modeling for tuple structure prediction.
By combining and permuting the sentiment elements, our multi-tasking model substantially outperforms task-specific models on a variety of ABSA tasks.
Detailed experiments show that our method significantly advances the state-of-the-art on benchmark datasets, in both supervised and low-resource settings.
We hope our research will shed light on generative tuple prediction.

\section*{Limitations}

Despite the state-of-the-art performances, our proposed methods still have some limitations for future directions.
\textbf{Firstly}, multi-view prompting creates overheads of training and inference proportional to the number of views. For efficiency in practice, according to Figure \ref{fig:num-view}, \mvp~with a relatively small number of views behaves decently (e.g., 5 or 7).
\textbf{Secondly}, we apply a simple yet effective aggregation strategy to combine the results of multiple views. More advanced strategies can be explored.
\textbf{Lastly}, experiments only verified the consistent improvement on ABSA tasks, while intuitively, the idea of \mvp~that leverages multiple views can be expanded to any structure prediction tasks, such as information extraction, emotion-cause pair extraction, and stance detection.

\section*{Ethics Statement}
We conduct all the experiments on existing datasets widely used in previous public scientific papers. We keep fair and honest in our analysis of experimental results, and our work does not harm anyone. We open-sourced our code for further explorations.

As for the broader impact, this work may foster further research in sentiment analysis using generative methods, contributing to the simplification and automation of user opinion mining in reality. Nevertheless, this work fine-tunes large pre-trained language models to generate sentiment tuples. Due to the large pre-training corpus based on the Internet, the predicted sentiment polarity is subject to unexpected bias with respect to gender, race, and intersectional identities \cite{tan2019assessing}, which needs to be considered more broadly in the field of natural language processing.

\section*{Acknowledgements}

We express our gratitude to Jiayi Li and Chengze Yu for their detailed feedback on a draft of the paper. We also thank Junjie Wang and Taiqiang Wu for their helpful discussion on our presentation.
This work was partly supported by the National Key Research and Development Program of China (No. 2020YFB1708200),  the "Graph Neural Network Project" of Ping An Technology (Shenzhen) Co., Ltd. and AMiner.Shenzhen SciBrain fund.

\bibliography{anthology,custom}

\begin{thebibliography}{56}
\expandafter\ifx\csname natexlab\endcsname\relax\def\natexlab#1{#1}\fi

\bibitem[{Bao et~al.(2022)Bao, Wang, Jiang, Xiao, and
  Li}]{DBLP:conf/ijcai/BaoWJXL22}
Xiaoyi Bao, Zhongqing Wang, Xiaotong Jiang, Rong Xiao, and Shoushan Li. 2022.
\newblock \href {https://doi.org/10.24963/ijcai.2022/561} {Aspect-based
  sentiment analysis with opinion tree generation}.
\newblock In \emph{Proceedings of the Thirty-First International Joint
  Conference on Artificial Intelligence, {IJCAI} 2022, Vienna, Austria, 23-29
  July 2022}, pages 4044--4050. ijcai.org.

\bibitem[{Bu et~al.(2021)Bu, Ren, Zheng, Yang, Wang, Zhang, and
  Wu}]{bu-etal-2021-asap}
Jiahao Bu, Lei Ren, Shuang Zheng, Yang Yang, Jingang Wang, Fuzheng Zhang, and
  Wei Wu. 2021.
\newblock \href {https://doi.org/10.18653/v1/2021.naacl-main.167} {{ASAP}: A
  {C}hinese review dataset towards aspect category sentiment analysis and
  rating prediction}.
\newblock In \emph{Proceedings of the 2021 Conference of the North American
  Chapter of the Association for Computational Linguistics: Human Language
  Technologies}, pages 2069--2079, Online. Association for Computational
  Linguistics.

\bibitem[{Cai et~al.(2020{\natexlab{a}})Cai, Tu, Zhou, Yu, and
  Xia}]{DBLP:conf/coling/CaiTZYX20}
Hongjie Cai, Yaofeng Tu, Xiangsheng Zhou, Jianfei Yu, and Rui Xia.
  2020{\natexlab{a}}.
\newblock \href {https://doi.org/10.18653/v1/2020.coling-main.72}
  {Aspect-category based sentiment analysis with hierarchical graph
  convolutional network}.
\newblock In \emph{Proceedings of the 28th International Conference on
  Computational Linguistics, {COLING} 2020, Barcelona, Spain (Online), December
  8-13, 2020}, pages 833--843. International Committee on Computational
  Linguistics.

\bibitem[{Cai et~al.(2020{\natexlab{b}})Cai, Tu, Zhou, Yu, and
  Xia}]{cai-etal-2020-aspect}
Hongjie Cai, Yaofeng Tu, Xiangsheng Zhou, Jianfei Yu, and Rui Xia.
  2020{\natexlab{b}}.
\newblock \href {https://doi.org/10.18653/v1/2020.coling-main.72}
  {Aspect-category based sentiment analysis with hierarchical graph
  convolutional network}.
\newblock In \emph{Proceedings of the 28th International Conference on
  Computational Linguistics}, pages 833--843, Barcelona, Spain (Online).
  International Committee on Computational Linguistics.

\bibitem[{Cai et~al.(2021)Cai, Xia, and Yu}]{cai-etal-2021-aspect}
Hongjie Cai, Rui Xia, and Jianfei Yu. 2021.
\newblock \href {https://doi.org/10.18653/v1/2021.acl-long.29}
  {Aspect-category-opinion-sentiment quadruple extraction with implicit aspects
  and opinions}.
\newblock In \emph{Proceedings of the 59th Annual Meeting of the Association
  for Computational Linguistics and the 11th International Joint Conference on
  Natural Language Processing (Volume 1: Long Papers)}, pages 340--350, Online.
  Association for Computational Linguistics.

\bibitem[{Cao et~al.(2021)Cao, Izacard, Riedel, and
  Petroni}]{DBLP:conf/iclr/CaoI0P21}
Nicola~De Cao, Gautier Izacard, Sebastian Riedel, and Fabio Petroni. 2021.
\newblock \href {https://openreview.net/forum?id=5k8F6UU39V} {Autoregressive
  entity retrieval}.
\newblock In \emph{9th International Conference on Learning Representations,
  {ICLR} 2021, Virtual Event, Austria, May 3-7, 2021}. OpenReview.net.

\bibitem[{Chen et~al.(2022)Chen, Zhai, Feng, Li, and
  Wang}]{chen-etal-2022-enhanced}
Hao Chen, Zepeng Zhai, Fangxiang Feng, Ruifan Li, and Xiaojie Wang. 2022.
\newblock \href {https://doi.org/10.18653/v1/2022.acl-long.212} {Enhanced
  multi-channel graph convolutional network for aspect sentiment triplet
  extraction}.
\newblock In \emph{Proceedings of the 60th Annual Meeting of the Association
  for Computational Linguistics (Volume 1: Long Papers)}, pages 2974--2985,
  Dublin, Ireland. Association for Computational Linguistics.

\bibitem[{Chen et~al.(2017)Chen, Sun, Bing, and
  Yang}]{chen-etal-2017-recurrent}
Peng Chen, Zhongqian Sun, Lidong Bing, and Wei Yang. 2017.
\newblock \href {https://doi.org/10.18653/v1/D17-1047} {Recurrent attention
  network on memory for aspect sentiment analysis}.
\newblock In \emph{Proceedings of the 2017 Conference on Empirical Methods in
  Natural Language Processing}, pages 452--461, Copenhagen, Denmark.
  Association for Computational Linguistics.

\bibitem[{Chen et~al.(2020)Chen, Liu, Wang, Zhang, and
  Chi}]{chen-etal-2020-synchronous}
Shaowei Chen, Jie Liu, Yu~Wang, Wenzheng Zhang, and Ziming Chi. 2020.
\newblock \href {https://doi.org/10.18653/v1/2020.acl-main.582} {Synchronous
  double-channel recurrent network for aspect-opinion pair extraction}.
\newblock In \emph{Proceedings of the 58th Annual Meeting of the Association
  for Computational Linguistics}, pages 6515--6524, Online. Association for
  Computational Linguistics.

\bibitem[{Chen and Qian(2020)}]{chen-qian-2020-relation}
Zhuang Chen and Tieyun Qian. 2020.
\newblock \href {https://doi.org/10.18653/v1/2020.acl-main.340} {Relation-aware
  collaborative learning for unified aspect-based sentiment analysis}.
\newblock In \emph{Proceedings of the 58th Annual Meeting of the Association
  for Computational Linguistics}, pages 3685--3694, Online. Association for
  Computational Linguistics.

\bibitem[{Gao et~al.(2022)Gao, Fang, Liu, Liu, Liu, Liu, Bao, and
  Yan}]{gao-etal-2022-lego}
Tianhao Gao, Jun Fang, Hanyu Liu, Zhiyuan Liu, Chao Liu, Pengzhang Liu, Yongjun
  Bao, and Weipeng Yan. 2022.
\newblock \href {https://aclanthology.org/2022.coling-1.610} {{LEGO}-{ABSA}: A
  prompt-based task assemblable unified generative framework for multi-task
  aspect-based sentiment analysis}.
\newblock In \emph{Proceedings of the 29th International Conference on
  Computational Linguistics}, pages 7002--7012, Gyeongju, Republic of Korea.
  International Committee on Computational Linguistics.

\bibitem[{Hu et~al.(2022)Hu, Wu, Gao, Bai, and
  Zhao}]{hu-etal-2022-improving-aspect}
Mengting Hu, Yike Wu, Hang Gao, Yinhao Bai, and Shiwan Zhao. 2022.
\newblock \href {https://aclanthology.org/2022.emnlp-main.538} {Improving
  aspect sentiment quad prediction via template-order data augmentation}.
\newblock In \emph{Proceedings of the 2022 Conference on Empirical Methods in
  Natural Language Processing}, pages 7889--7900, Abu Dhabi, United Arab
  Emirates. Association for Computational Linguistics.

\bibitem[{Huang and Carley(2018)}]{huang-carley-2018-parameterized}
Binxuan Huang and Kathleen Carley. 2018.
\newblock \href {https://doi.org/10.18653/v1/D18-1136} {Parameterized
  convolutional neural networks for aspect level sentiment classification}.
\newblock In \emph{Proceedings of the 2018 Conference on Empirical Methods in
  Natural Language Processing}, pages 1091--1096, Brussels, Belgium.
  Association for Computational Linguistics.

\bibitem[{Lei et~al.(2018)Lei, Yang, Yang, and Liu}]{lei-etal-2018-multi}
Zeyang Lei, Yujiu Yang, Min Yang, and Yi~Liu. 2018.
\newblock \href {https://doi.org/10.18653/v1/P18-2120} {A
  multi-sentiment-resource enhanced attention network for sentiment
  classification}.
\newblock In \emph{Proceedings of the 56th Annual Meeting of the Association
  for Computational Linguistics (Volume 2: Short Papers)}, pages 758--763,
  Melbourne, Australia. Association for Computational Linguistics.

\bibitem[{Lei et~al.(2019)Lei, Yang, Yang, Zhao, Guo, and
  Liu}]{DBLP:conf/aaai/LeiYYZG019}
Zeyang Lei, Yujiu Yang, Min Yang, Wei Zhao, Jun Guo, and Yi~Liu. 2019.
\newblock \href {https://doi.org/10.1609/aaai.v33i01.33016650} {A human-like
  semantic cognition network for aspect-level sentiment classification}.
\newblock In \emph{The Thirty-Third {AAAI} Conference on Artificial
  Intelligence, {AAAI} 2019, The Thirty-First Innovative Applications of
  Artificial Intelligence Conference, {IAAI} 2019, The Ninth {AAAI} Symposium
  on Educational Advances in Artificial Intelligence, {EAAI} 2019, Honolulu,
  Hawaii, USA, January 27 - February 1, 2019}, pages 6650--6657. {AAAI} Press.

\bibitem[{Lewis et~al.(2020)Lewis, Liu, Goyal, Ghazvininejad, Mohamed, Levy,
  Stoyanov, and Zettlemoyer}]{lewis-etal-2020-bart}
Mike Lewis, Yinhan Liu, Naman Goyal, Marjan Ghazvininejad, Abdelrahman Mohamed,
  Omer Levy, Veselin Stoyanov, and Luke Zettlemoyer. 2020.
\newblock \href {https://doi.org/10.18653/v1/2020.acl-main.703} {{BART}:
  Denoising sequence-to-sequence pre-training for natural language generation,
  translation, and comprehension}.
\newblock In \emph{Proceedings of the 58th Annual Meeting of the Association
  for Computational Linguistics}, pages 7871--7880, Online. Association for
  Computational Linguistics.

\bibitem[{Liu et~al.(2021{\natexlab{a}})Liu, Teng, Cui, Liu, and
  Zhang}]{liu-etal-2021-solving}
Jian Liu, Zhiyang Teng, Leyang Cui, Hanmeng Liu, and Yue Zhang.
  2021{\natexlab{a}}.
\newblock \href {https://doi.org/10.18653/v1/2021.emnlp-main.361} {Solving
  aspect category sentiment analysis as a text generation task}.
\newblock In \emph{Proceedings of the 2021 Conference on Empirical Methods in
  Natural Language Processing}, pages 4406--4416, Online and Punta Cana,
  Dominican Republic. Association for Computational Linguistics.

\bibitem[{Liu et~al.(2015)Liu, Joty, and Meng}]{liu-etal-2015-fine}
Pengfei Liu, Shafiq Joty, and Helen Meng. 2015.
\newblock \href {https://doi.org/10.18653/v1/D15-1168} {Fine-grained opinion
  mining with recurrent neural networks and word embeddings}.
\newblock In \emph{Proceedings of the 2015 Conference on Empirical Methods in
  Natural Language Processing}, pages 1433--1443, Lisbon, Portugal. Association
  for Computational Linguistics.

\bibitem[{Liu et~al.(2021{\natexlab{b}})Liu, Yuan, Fu, Jiang, Hayashi, and
  Neubig}]{DBLP:journals/corr/abs-2107-13586}
Pengfei Liu, Weizhe Yuan, Jinlan Fu, Zhengbao Jiang, Hiroaki Hayashi, and
  Graham Neubig. 2021{\natexlab{b}}.
\newblock \href {http://arxiv.org/abs/2107.13586} {Pre-train, prompt, and
  predict: {A} systematic survey of prompting methods in natural language
  processing}.
\newblock \emph{CoRR}, abs/2107.13586.

\bibitem[{Loshchilov and Hutter(2019)}]{DBLP:conf/iclr/LoshchilovH19}
Ilya Loshchilov and Frank Hutter. 2019.
\newblock \href {https://openreview.net/forum?id=Bkg6RiCqY7} {Decoupled weight
  decay regularization}.
\newblock In \emph{7th International Conference on Learning Representations,
  {ICLR} 2019, New Orleans, LA, USA, May 6-9, 2019}. OpenReview.net.

\bibitem[{Lu et~al.(2022{\natexlab{a}})Lu, Bartolo, Moore, Riedel, and
  Stenetorp}]{lu-etal-2022-fantastically}
Yao Lu, Max Bartolo, Alastair Moore, Sebastian Riedel, and Pontus Stenetorp.
  2022{\natexlab{a}}.
\newblock \href {https://doi.org/10.18653/v1/2022.acl-long.556} {Fantastically
  ordered prompts and where to find them: Overcoming few-shot prompt order
  sensitivity}.
\newblock In \emph{Proceedings of the 60th Annual Meeting of the Association
  for Computational Linguistics (Volume 1: Long Papers)}, pages 8086--8098,
  Dublin, Ireland. Association for Computational Linguistics.

\bibitem[{Lu et~al.(2022{\natexlab{b}})Lu, Liu, Dai, Xiao, Lin, Han, Sun, and
  Wu}]{lu-etal-2022-unified}
Yaojie Lu, Qing Liu, Dai Dai, Xinyan Xiao, Hongyu Lin, Xianpei Han, Le~Sun, and
  Hua Wu. 2022{\natexlab{b}}.
\newblock \href {https://doi.org/10.18653/v1/2022.acl-long.395} {Unified
  structure generation for universal information extraction}.
\newblock In \emph{Proceedings of the 60th Annual Meeting of the Association
  for Computational Linguistics (Volume 1: Long Papers)}, pages 5755--5772,
  Dublin, Ireland. Association for Computational Linguistics.

\bibitem[{Luo et~al.(2019)Luo, Li, Liu, and Zhang}]{luo-etal-2019-doer}
Huaishao Luo, Tianrui Li, Bing Liu, and Junbo Zhang. 2019.
\newblock \href {https://doi.org/10.18653/v1/P19-1056} {{DOER}: Dual
  cross-shared {RNN} for aspect term-polarity co-extraction}.
\newblock In \emph{Proceedings of the 57th Annual Meeting of the Association
  for Computational Linguistics}, pages 591--601, Florence, Italy. Association
  for Computational Linguistics.

\bibitem[{Ma et~al.(2019)Ma, Li, Wu, Xie, and Wang}]{ma-etal-2019-exploring}
Dehong Ma, Sujian Li, Fangzhao Wu, Xing Xie, and Houfeng Wang. 2019.
\newblock \href {https://doi.org/10.18653/v1/P19-1344} {Exploring
  sequence-to-sequence learning in aspect term extraction}.
\newblock In \emph{Proceedings of the 57th Annual Meeting of the Association
  for Computational Linguistics}, pages 3538--3547, Florence, Italy.
  Association for Computational Linguistics.

\bibitem[{Mao et~al.(2022)Mao, Shen, Yang, Zhu, and
  Cai}]{mao-etal-2022-seq2path}
Yue Mao, Yi~Shen, Jingchao Yang, Xiaoying Zhu, and Longjun Cai. 2022.
\newblock \href {https://doi.org/10.18653/v1/2022.findings-acl.174}
  {{S}eq2{P}ath: Generating sentiment tuples as paths of a tree}.
\newblock In \emph{Findings of the Association for Computational Linguistics:
  ACL 2022}, pages 2215--2225, Dublin, Ireland. Association for Computational
  Linguistics.

\bibitem[{Mishra et~al.(2022)Mishra, Khashabi, Baral, and
  Hajishirzi}]{mishra-etal-2022-cross}
Swaroop Mishra, Daniel Khashabi, Chitta Baral, and Hannaneh Hajishirzi. 2022.
\newblock \href {https://doi.org/10.18653/v1/2022.acl-long.244} {Cross-task
  generalization via natural language crowdsourcing instructions}.
\newblock In \emph{Proceedings of the 60th Annual Meeting of the Association
  for Computational Linguistics (Volume 1: Long Papers)}, pages 3470--3487,
  Dublin, Ireland. Association for Computational Linguistics.

\bibitem[{Paolini et~al.(2021)Paolini, Athiwaratkun, Krone, Ma, Achille,
  Anubhai, dos Santos, Xiang, and Soatto}]{DBLP:conf/iclr/PaoliniAKMAASXS21}
Giovanni Paolini, Ben Athiwaratkun, Jason Krone, Jie Ma, Alessandro Achille,
  Rishita Anubhai, C{\'{\i}}cero~Nogueira dos Santos, Bing Xiang, and Stefano
  Soatto. 2021.
\newblock \href {https://openreview.net/forum?id=US-TP-xnXI} {Structured
  prediction as translation between augmented natural languages}.
\newblock In \emph{9th International Conference on Learning Representations,
  {ICLR} 2021, Virtual Event, Austria, May 3-7, 2021}. OpenReview.net.

\bibitem[{Peng et~al.(2020)Peng, Xu, Bing, Huang, Lu, and
  Si}]{DBLP:conf/aaai/PengXBHLS20}
Haiyun Peng, Lu~Xu, Lidong Bing, Fei Huang, Wei Lu, and Luo Si. 2020.
\newblock \href {https://ojs.aaai.org/index.php/AAAI/article/view/6383}
  {Knowing what, how and why: {A} near complete solution for aspect-based
  sentiment analysis}.
\newblock In \emph{The Thirty-Fourth {AAAI} Conference on Artificial
  Intelligence, {AAAI} 2020, The Thirty-Second Innovative Applications of
  Artificial Intelligence Conference, {IAAI} 2020, The Tenth {AAAI} Symposium
  on Educational Advances in Artificial Intelligence, {EAAI} 2020, New York,
  NY, USA, February 7-12, 2020}, pages 8600--8607. {AAAI} Press.

\bibitem[{Phan and Ogunbona(2020)}]{phan-ogunbona-2020-modelling}
Minh~Hieu Phan and Philip~O. Ogunbona. 2020.
\newblock \href {https://doi.org/10.18653/v1/2020.acl-main.293} {Modelling
  context and syntactical features for aspect-based sentiment analysis}.
\newblock In \emph{Proceedings of the 58th Annual Meeting of the Association
  for Computational Linguistics}, pages 3211--3220, Online. Association for
  Computational Linguistics.

\bibitem[{Pontiki et~al.(2016)Pontiki, Galanis, Papageorgiou, Androutsopoulos,
  Manandhar, AL-Smadi, Al-Ayyoub, Zhao, Qin, De~Clercq, Hoste, Apidianaki,
  Tannier, Loukachevitch, Kotelnikov, Bel, Jim{\'e}nez-Zafra, and
  Eryi{\u{g}}it}]{pontiki-etal-2016-semeval}
Maria Pontiki, Dimitris Galanis, Haris Papageorgiou, Ion Androutsopoulos,
  Suresh Manandhar, Mohammad AL-Smadi, Mahmoud Al-Ayyoub, Yanyan Zhao, Bing
  Qin, Orph{\'e}e De~Clercq, V{\'e}ronique Hoste, Marianna Apidianaki, Xavier
  Tannier, Natalia Loukachevitch, Evgeniy Kotelnikov, Nuria Bel,
  Salud~Mar{\'\i}a Jim{\'e}nez-Zafra, and G{\"u}l{\c{s}}en Eryi{\u{g}}it. 2016.
\newblock \href {https://doi.org/10.18653/v1/S16-1002} {{S}em{E}val-2016 task
  5: Aspect based sentiment analysis}.
\newblock In \emph{Proceedings of the 10th International Workshop on Semantic
  Evaluation ({S}em{E}val-2016)}, pages 19--30, San Diego, California.
  Association for Computational Linguistics.

\bibitem[{Pontiki et~al.(2015)Pontiki, Galanis, Papageorgiou, Manandhar, and
  Androutsopoulos}]{pontiki-etal-2015-semeval}
Maria Pontiki, Dimitris Galanis, Haris Papageorgiou, Suresh Manandhar, and Ion
  Androutsopoulos. 2015.
\newblock \href {https://doi.org/10.18653/v1/S15-2082} {{S}em{E}val-2015 task
  12: Aspect based sentiment analysis}.
\newblock In \emph{Proceedings of the 9th International Workshop on Semantic
  Evaluation ({S}em{E}val 2015)}, pages 486--495, Denver, Colorado. Association
  for Computational Linguistics.

\bibitem[{Qiu et~al.(2011)Qiu, Liu, Bu, and Chen}]{qiu-etal-2011-opinion}
Guang Qiu, Bing Liu, Jiajun Bu, and Chun Chen. 2011.
\newblock \href {https://doi.org/10.1162/coli_a_00034} {Opinion word expansion
  and target extraction through double propagation}.
\newblock \emph{Computational Linguistics}, 37(1):9--27.

\bibitem[{Raffel et~al.(2020)Raffel, Shazeer, Roberts, Lee, Narang, Matena,
  Zhou, Li, and Liu}]{DBLP:journals/jmlr/RaffelSRLNMZLL20}
Colin Raffel, Noam Shazeer, Adam Roberts, Katherine Lee, Sharan Narang, Michael
  Matena, Yanqi Zhou, Wei Li, and Peter~J. Liu. 2020.
\newblock \href {http://jmlr.org/papers/v21/20-074.html} {Exploring the limits
  of transfer learning with a unified text-to-text transformer}.
\newblock \emph{J. Mach. Learn. Res.}, 21:140:1--140:67.

\bibitem[{Stanovich and West(2000)}]{stanovich2000individual}
Keith~E Stanovich and Richard~F West. 2000.
\newblock Individual differences in reasoning: Implications for the rationality
  debate?
\newblock \emph{Behavioral and brain sciences}, 23(5):645--665.

\bibitem[{Sun et~al.(2022)Sun, Liu, Qiu, and
  Huang}]{DBLP:journals/ijautcomp/SunLQH22}
Tianxiang Sun, Xiangyang Liu, Xipeng Qiu, and Xuan{-}Jing Huang. 2022.
\newblock \href {https://doi.org/10.1007/s11633-022-1331-6} {Paradigm shift in
  natural language processing}.
\newblock \emph{Int. J. Autom. Comput.}, 19(3):169--183.

\bibitem[{Tan and Celis(2019)}]{tan2019assessing}
Yi~Chern Tan and L~Elisa Celis. 2019.
\newblock Assessing social and intersectional biases in contextualized word
  representations.
\newblock \emph{Advances in Neural Information Processing Systems}, 32.

\bibitem[{Vaswani et~al.(2017)Vaswani, Shazeer, Parmar, Uszkoreit, Jones,
  Gomez, Kaiser, and Polosukhin}]{DBLP:conf/nips/VaswaniSPUJGKP17}
Ashish Vaswani, Noam Shazeer, Niki Parmar, Jakob Uszkoreit, Llion Jones,
  Aidan~N. Gomez, Lukasz Kaiser, and Illia Polosukhin. 2017.
\newblock \href
  {https://proceedings.neurips.cc/paper/2017/hash/3f5ee243547dee91fbd053c1c4a845aa-Abstract.html}
  {Attention is all you need}.
\newblock In \emph{Advances in Neural Information Processing Systems 30: Annual
  Conference on Neural Information Processing Systems 2017, December 4-9, 2017,
  Long Beach, CA, {USA}}, pages 5998--6008.

\bibitem[{Wan et~al.(2020)Wan, Yang, Du, Liu, Qi, and
  Pan}]{DBLP:conf/aaai/WanYDLQP20}
Hai Wan, Yufei Yang, Jianfeng Du, Yanan Liu, Kunxun Qi, and Jeff~Z. Pan. 2020.
\newblock \href {https://ojs.aaai.org/index.php/AAAI/article/view/6447}
  {Target-aspect-sentiment joint detection for aspect-based sentiment
  analysis}.
\newblock In \emph{The Thirty-Fourth {AAAI} Conference on Artificial
  Intelligence, {AAAI} 2020, The Thirty-Second Innovative Applications of
  Artificial Intelligence Conference, {IAAI} 2020, The Tenth {AAAI} Symposium
  on Educational Advances in Artificial Intelligence, {EAAI} 2020, New York,
  NY, USA, February 7-12, 2020}, pages 9122--9129. {AAAI} Press.

\bibitem[{Wang et~al.(2017)Wang, Pan, Dahlmeier, and
  Xiao}]{DBLP:conf/aaai/WangPDX17}
Wenya Wang, Sinno~Jialin Pan, Daniel Dahlmeier, and Xiaokui Xiao. 2017.
\newblock \href {http://aaai.org/ocs/index.php/AAAI/AAAI17/paper/view/14441}
  {Coupled multi-layer attentions for co-extraction of aspect and opinion
  terms}.
\newblock In \emph{Proceedings of the Thirty-First {AAAI} Conference on
  Artificial Intelligence, February 4-9, 2017, San Francisco, California,
  {USA}}, pages 3316--3322. {AAAI} Press.

\bibitem[{Wang et~al.(2022{\natexlab{a}})Wang, Wei, Schuurmans, Le, Chi, and
  Zhou}]{DBLP:journals/corr/abs-2207-00747}
Xuezhi Wang, Jason Wei, Dale Schuurmans, Quoc~V. Le, Ed~H. Chi, and Denny Zhou.
  2022{\natexlab{a}}.
\newblock \href {https://doi.org/10.48550/arXiv.2207.00747}
  {Rationale-augmented ensembles in language models}.
\newblock \emph{CoRR}, abs/2207.00747.

\bibitem[{Wang et~al.(2022{\natexlab{b}})Wang, Wei, Schuurmans, Le, Chi, and
  Zhou}]{DBLP:journals/corr/abs-2203-11171}
Xuezhi Wang, Jason Wei, Dale Schuurmans, Quoc~V. Le, Ed~H. Chi, and Denny Zhou.
  2022{\natexlab{b}}.
\newblock \href {https://doi.org/10.48550/arXiv.2203.11171} {Self-consistency
  improves chain of thought reasoning in language models}.
\newblock \emph{CoRR}, abs/2203.11171.

\bibitem[{Wang et~al.(2016)Wang, Huang, Zhu, and
  Zhao}]{wang-etal-2016-attention}
Yequan Wang, Minlie Huang, Xiaoyan Zhu, and Li~Zhao. 2016.
\newblock \href {https://doi.org/10.18653/v1/D16-1058} {Attention-based {LSTM}
  for aspect-level sentiment classification}.
\newblock In \emph{Proceedings of the 2016 Conference on Empirical Methods in
  Natural Language Processing}, pages 606--615, Austin, Texas. Association for
  Computational Linguistics.

\bibitem[{Wang et~al.(2022{\natexlab{c}})Wang, Xia, and
  Yu}]{DBLP:journals/corr/abs-2211-10986}
Zengzhi Wang, Rui Xia, and Jianfei Yu. 2022{\natexlab{c}}.
\newblock \href {https://doi.org/10.48550/arXiv.2211.10986} {Unifiedabsa: {A}
  unified {ABSA} framework based on multi-task instruction tuning}.
\newblock \emph{CoRR}, abs/2211.10986.

\bibitem[{Wei et~al.(2022{\natexlab{a}})Wei, Bosma, Zhao, Guu, Yu, Lester, Du,
  Dai, and Le}]{DBLP:conf/iclr/WeiBZGYLDDL22}
Jason Wei, Maarten Bosma, Vincent~Y. Zhao, Kelvin Guu, Adams~Wei Yu, Brian
  Lester, Nan Du, Andrew~M. Dai, and Quoc~V. Le. 2022{\natexlab{a}}.
\newblock \href {https://openreview.net/forum?id=gEZrGCozdqR} {Finetuned
  language models are zero-shot learners}.
\newblock In \emph{The Tenth International Conference on Learning
  Representations, {ICLR} 2022, Virtual Event, April 25-29, 2022}.
  OpenReview.net.

\bibitem[{Wei et~al.(2022{\natexlab{b}})Wei, Wang, Schuurmans, Bosma, Chi, Le,
  and Zhou}]{DBLP:journals/corr/abs-2201-11903}
Jason Wei, Xuezhi Wang, Dale Schuurmans, Maarten Bosma, Ed~H. Chi, Quoc Le, and
  Denny Zhou. 2022{\natexlab{b}}.
\newblock \href {http://arxiv.org/abs/2201.11903} {Chain of thought prompting
  elicits reasoning in large language models}.
\newblock \emph{Conference on Neural Information Processing Systems (NeurIPS)}.

\bibitem[{Wolf et~al.(2020)Wolf, Debut, Sanh, Chaumond, Delangue, Moi, Cistac,
  Rault, Louf, Funtowicz, Davison, Shleifer, von Platen, Ma, Jernite, Plu, Xu,
  Le~Scao, Gugger, Drame, Lhoest, and Rush}]{wolf-etal-2020-transformers}
Thomas Wolf, Lysandre Debut, Victor Sanh, Julien Chaumond, Clement Delangue,
  Anthony Moi, Pierric Cistac, Tim Rault, Remi Louf, Morgan Funtowicz, Joe
  Davison, Sam Shleifer, Patrick von Platen, Clara Ma, Yacine Jernite, Julien
  Plu, Canwen Xu, Teven Le~Scao, Sylvain Gugger, Mariama Drame, Quentin Lhoest,
  and Alexander Rush. 2020.
\newblock \href {https://doi.org/10.18653/v1/2020.emnlp-demos.6} {Transformers:
  State-of-the-art natural language processing}.
\newblock In \emph{Proceedings of the 2020 Conference on Empirical Methods in
  Natural Language Processing: System Demonstrations}, pages 38--45, Online.
  Association for Computational Linguistics.

\bibitem[{Xu et~al.(2020)Xu, Li, Lu, and Bing}]{xu-etal-2020-position}
Lu~Xu, Hao Li, Wei Lu, and Lidong Bing. 2020.
\newblock \href {https://doi.org/10.18653/v1/2020.emnlp-main.183}
  {Position-aware tagging for aspect sentiment triplet extraction}.
\newblock In \emph{Proceedings of the 2020 Conference on Empirical Methods in
  Natural Language Processing (EMNLP)}, pages 2339--2349, Online. Association
  for Computational Linguistics.

\bibitem[{Yan et~al.(2021)Yan, Dai, Ji, Qiu, and Zhang}]{yan-etal-2021-unified}
Hang Yan, Junqi Dai, Tuo Ji, Xipeng Qiu, and Zheng Zhang. 2021.
\newblock \href {https://doi.org/10.18653/v1/2021.acl-long.188} {A unified
  generative framework for aspect-based sentiment analysis}.
\newblock In \emph{Proceedings of the 59th Annual Meeting of the Association
  for Computational Linguistics and the 11th International Joint Conference on
  Natural Language Processing (Volume 1: Long Papers)}, pages 2416--2429,
  Online. Association for Computational Linguistics.

\bibitem[{Yu et~al.(2023)Yu, Wu, Li, Bai, and Yang}]{yu2023syngen}
Chengze Yu, Taiqiang Wu, Jiayi Li, Xingyu Bai, and Yujiu Yang. 2023.
\newblock Syngen: A syntactic plug-and-play module for generative aspect-based
  sentiment analysis.
\newblock In \emph{ICASSP 2023-2023 IEEE International Conference on Acoustics,
  Speech and Signal Processing (ICASSP)}, pages 1--5. IEEE.

\bibitem[{Zhai et~al.(2022)Zhai, Chen, Feng, Li, and Wang}]{zhai-etal-2022-com}
Zepeng Zhai, Hao Chen, Fangxiang Feng, Ruifan Li, and Xiaojie Wang. 2022.
\newblock \href {https://aclanthology.org/2022.emnlp-main.212} {{COM}-{MRC}: A
  {CO}ntext-masked machine reading comprehension framework for aspect sentiment
  triplet extraction}.
\newblock In \emph{Proceedings of the 2022 Conference on Empirical Methods in
  Natural Language Processing}, pages 3230--3241, Abu Dhabi, United Arab
  Emirates. Association for Computational Linguistics.

\bibitem[{Zhang et~al.(2021{\natexlab{a}})Zhang, Deng, Li, Bing, and
  Lam}]{zhang-etal-2021-aspect}
Wenxuan Zhang, Yang Deng, Xin Li, Lidong Bing, and Wai Lam. 2021{\natexlab{a}}.
\newblock \href {https://doi.org/10.18653/v1/2021.findings-emnlp.390}
  {Aspect-based sentiment analysis in question answering forums}.
\newblock In \emph{Findings of the Association for Computational Linguistics:
  EMNLP 2021}, pages 4582--4591, Punta Cana, Dominican Republic. Association
  for Computational Linguistics.

\bibitem[{Zhang et~al.(2021{\natexlab{b}})Zhang, Deng, Li, Yuan, Bing, and
  Lam}]{zhang-etal-2021-aspect-sentiment}
Wenxuan Zhang, Yang Deng, Xin Li, Yifei Yuan, Lidong Bing, and Wai Lam.
  2021{\natexlab{b}}.
\newblock \href {https://doi.org/10.18653/v1/2021.emnlp-main.726} {Aspect
  sentiment quad prediction as paraphrase generation}.
\newblock In \emph{Proceedings of the 2021 Conference on Empirical Methods in
  Natural Language Processing}, pages 9209--9219, Online and Punta Cana,
  Dominican Republic. Association for Computational Linguistics.

\bibitem[{Zhang et~al.(2021{\natexlab{c}})Zhang, Li, Deng, Bing, and
  Lam}]{zhang-etal-2021-towards-generative}
Wenxuan Zhang, Xin Li, Yang Deng, Lidong Bing, and Wai Lam. 2021{\natexlab{c}}.
\newblock \href {https://doi.org/10.18653/v1/2021.acl-short.64} {Towards
  generative aspect-based sentiment analysis}.
\newblock In \emph{Proceedings of the 59th Annual Meeting of the Association
  for Computational Linguistics and the 11th International Joint Conference on
  Natural Language Processing (Volume 2: Short Papers)}, pages 504--510,
  Online. Association for Computational Linguistics.

\bibitem[{Zhang et~al.(2022)Zhang, Li, Deng, Bing, and
  Lam}]{DBLP:journals/corr/abs-2203-01054}
Wenxuan Zhang, Xin Li, Yang Deng, Lidong Bing, and Wai Lam. 2022.
\newblock \href {https://doi.org/10.48550/arXiv.2203.01054} {A survey on
  aspect-based sentiment analysis: Tasks, methods, and challenges}.
\newblock \emph{CoRR}, abs/2203.01054.

\bibitem[{Zhang et~al.(2021{\natexlab{d}})Zhang, Guo, and
  Kordjamshidi}]{zhang-etal-2021-towards}
Yue Zhang, Quan Guo, and Parisa Kordjamshidi. 2021{\natexlab{d}}.
\newblock \href {https://doi.org/10.18653/v1/2021.splurobonlp-1.5} {Towards
  navigation by reasoning over spatial configurations}.
\newblock In \emph{Proceedings of Second International Combined Workshop on
  Spatial Language Understanding and Grounded Communication for Robotics},
  pages 42--52, Online. Association for Computational Linguistics.

\bibitem[{Zhou et~al.(2015)Zhou, Wan, and Xiao}]{DBLP:conf/aaai/ZhouWX15}
Xinjie Zhou, Xiaojun Wan, and Jianguo Xiao. 2015.
\newblock \href {http://www.aaai.org/ocs/index.php/AAAI/AAAI15/paper/view/9764}
  {Representation learning for aspect category detection in online reviews}.
\newblock In \emph{Proceedings of the Twenty-Ninth {AAAI} Conference on
  Artificial Intelligence, January 25-30, 2015, Austin, Texas, {USA}}, pages
  417--424. {AAAI} Press.

\end{thebibliography}
\bibliographystyle{acl_natbib}

\newpage
\appendix

\section{Data Statistics}
\label{sec:appendix:data}

Table \ref{tab:data-sta-all} shows the data statistics of all datasets of the ASQP, ACOS, ASTE and TASD task. For fair comparison, we keep the same train/dev/test division as previous works.

\section{Constrained Decoding}
\label{sec:appendix:cd}
To make sure the predicted output complies with the mandatory format, we apply the constrained decoding (CD) algorithm in experiments. Rather than search the whole vocabulary for the next token to decode, which may make the model generate invalid sequences that do not match our expectations, CD adjusts the candidate list dynamically in terms of the current state token by token. If the current token is decoded as '[', which means the next token should be selected from a list of terms, i.e., $[\mathtt{A}]$, $[\mathtt{O}]$, $[\mathtt{S}]$ and $[\mathtt{C}]$. Additionally, CD tracks the current term and decodes the next following tokens based on Table \ref{tab:cd}.

\begin{table}[h]
  \centering
  \resizebox{\columnwidth}{!}{
\begin{tabular}{cr}
\toprule
Current Term & Candidate tokens                               \\ \hline
$[\mathtt{A}]$          & Input sentence, {$[\mathtt{SSEP}]$}      \\
$[\mathtt{O}]$           & Input sentence, {$[\mathtt{SSEP}]$}      \\
$[\mathtt{S}]$           & great, bad, neutral, {$[\mathtt{SSEP}]$} \\
$[\mathtt{C}]$          & All categories, {$[\mathtt{SSEP}]$}      \\ \bottomrule
\end{tabular}
}
\caption{Candidate lists of different terms}
\label{tab:cd}
\end{table}

\section{Detailed Experimental Settings}
\label{sec:appendix:exps}
Hyper-parameters for all experiments can be found in Table \ref{tab:param}. 
We employ the AdamW \cite{DBLP:conf/iclr/LoshchilovH19} as the optimizer. All experiments are carried out with an Nvidia RTX 3090 GPU.

\section{Comparison with ChatGPT}
\label{sec:appendix:chatgpt}

\subsection{Experiments}

We refined the prompt design \footnote{Designed based on \url{https://github.com/RidongHan/Evaluation-of-ChatGPT-on-Information-Extraction}.} and evaluated ChatGPT (\texttt{gpt-3.5-turbo}) on four ABSA tasks. Due to budget constraints, we tested it with 200 random samples for each task.

The experimental results, shown in Table \ref{tab:chatgpt-result}, highlight the remarkable performance advantage of cross-task transferred and fully supervised MvP compared to the few-shot prompted ChatGPT (+7.06\% and +22.84\% absolute F1 scores).

\begin{table*}[h]
\centering
\begin{tabular}{l|ccccc}
\toprule
\multirow{2}{*}{\textbf{Hyperparameters}} &
  \multicolumn{1}{c|}{\multirow{2}{*}{\mvp~(All supervised)}} &
  \multicolumn{4}{c}{\mvp~(Low Resource)} \\ \cline{3-6} 
 &
  \multicolumn{1}{c|}{} &
  \multicolumn{1}{c|}{1\%, 2\%, 3\%, 5\%} &
  \multicolumn{1}{c|}{10\%, 20\%} &
  \multicolumn{1}{c|}{30\%} &
  50\% \\ \hline
Epoch &
  \multicolumn{1}{c|}{20} &
  \multicolumn{1}{c|}{100} &
  \multicolumn{1}{c|}{50} &
  \multicolumn{1}{c|}{30} &
  20 \\ \hline
Batch Size    & \multicolumn{1}{c|}{16} & \multicolumn{4}{c}{8} \\ \hline
Learning Rate & \multicolumn{5}{c}{1e-4}                        \\
\bottomrule
\end{tabular}
\caption{Hyper-parameters for all supervised and low-resource settings}
\label{tab:param}
\end{table*}

\begin{table*}[ht]
\centering
\begin{tabular}{llcccc}
\toprule
\textbf{Methods} & \textbf{Data} & \textbf{ASQP (R15)} & \textbf{ACOS (Rest)} & \textbf{TASD (R16)} & \textbf{ASTE (L14)} \\
\midrule
ChatGPT & zero-shot & 22.87 & 27.11 & 34.08 & 36.05 \\
ChatGPT & few-shot & \underline{34.27} & 37.71 & 46.51 & 38.12 \\
\mvp~(transfer) & few-shot & 28.69 & \underline{39.24} & \underline{68.49} & \underline{48.43} \\
\mvp & full-data & \textbf{51.04} & \textbf{61.54} & \textbf{72.76} & \textbf{63.33} \\
\bottomrule
\end{tabular}
\caption{\label{tab:chatgpt-result}
Comparison with ChatGPT (\texttt{gpt-3.5-turbo}). F1 scores are reported. The best
results are in bold, while the second best are underlined. The few-shot results of \mvp~(transfer) are from Table \ref{tab:few}.}

\bigskip
\centering
\setlength{\tabcolsep}{4pt}
\begin{tabular}{l||cc|cc|cc|cccc||g}
\toprule
\multirow{2}{*}{\textbf{Methods}} & \multicolumn{2}{c|}{\textbf{ASQP}} & \multicolumn{2}{c|}{\textbf{ACOS}} & \multicolumn{2}{c|}{\textbf{TASD}} &  \multicolumn{4}{c||}{\textbf{ASTE}} & \cellcolor{lightgray}\ \\
               & \textbf{R15} & \textbf{R16} & \textbf{Lap} & \textbf{Rest} & \textbf{R15} & \textbf{R16} & \textbf{L14} & \textbf{R14} & \textbf{R15} & \textbf{R16} & \multirow{-2}{*}{\textbf{AVG}} \\
\midrule
SvP (random) & 48.32 & 58.94 & 43.61 & 58.16 & 63.42 & 71.60 & 62.36 & 71.64 & 62.31 & 71.59 & 61.19 \\
SvP (heuristic) & 49.02 & 59.56 & 43.83 & 59.38 & 61.98 & 71.57 & 62.09 & 72.61 & 65.29 & 73.27 & 61.86 \\
SvP (rank) & 48.39 & 58.67 & 43.86 & 59.57 & 62.93 & 71.26 & 62.83 & 72.71 & 63.57 & 71.79 & 61.56 \\
\mvp & \textbf{51.04} & \textbf{60.39} & \textbf{43.92} & \textbf{61.54} & \textbf{64.53} & \textbf{72.76} & \textbf{63.33} & \textbf{74.05} & \textbf{65.89} & \textbf{73.48} & \textbf{63.09} \\
\bottomrule
\end{tabular}
\caption{\label{tab:svp-result}
Additional comparison with single-view prompting on 10 datasets of ASQP, ACOS, TASD and ASTE tasks. F1 scores are reported. 
}
\end{table*}

\subsection{Prompts for ChatGPT}
We present zero- and few-shot prompts for ASQP (R15) in Listing \ref{lst:list1} and \ref{lst:list2}. For prompts related to other tasks, please refer to our released code.

\begin{figure*}
    \centering
\begin{lstlisting}[caption={Zero-shot Prompt for ASQP (R15).},basicstyle=\small, label=lst:list1]
According to the following sentiment elements definition: 

- The 'aspect term' refers to a specific feature, attribute, or aspect of a product or service that a user may express an opinion about, the aspect term might be 'null' for implicit aspect.
- The 'opinion term' refers to the sentiment or attitude expressed by a user towards a particular aspect or feature of a product or service, the aspect term might be 'null' for implicit opinion.
- The 'aspect category' refers to the category that aspect belongs to, and the available catgories includes: 'location general', 'food prices', 'food quality', 'food general', 'ambience general', 'service general', 'restaurant prices', 'drinks prices', 'restaurant miscellaneous', 'drinks quality', 'drinks style_options', 'restaurant general' and 'food style_options'.
- The 'sentiment polarity' refers to the degree of positivity, negativity or neutrality expressed in the opinion towards a particular aspect or feature of a product or service, and the available polarities inlcudes: 'positive', 'negative' and 'neutral'.

Recognize all sentiment elements with their corresponding aspect terms, aspect categories, opinion terms and sentiment polarity in the following text with the format of [('aspect term', 'opinion term', 'aspect category', 'sentiment polarity'), ...]: 
\end{lstlisting}
\end{figure*}

\begin{figure*}
    \centering
\begin{lstlisting}[caption={Few-shot Prompt (10 shots) for ASQP (R15).}, basicstyle=\small, label=lst:list2]
According to the following sentiment elements definition: 

- The 'aspect term' refers to a specific feature, attribute, or aspect of a product or service that a user may express an opinion about, the aspect term might be 'null' for implicit aspect.
- The 'opinion term' refers to the sentiment or attitude expressed by a user towards a particular aspect or feature of a product or service, the aspect term might be 'null' for implicit opinion.
- The 'aspect category' refers to the category that aspect belongs to, and the available catgories includes: 'location general', 'food prices', 'food quality', 'food general', 'ambience general', 'service general', 'restaurant prices', 'drinks prices', 'restaurant miscellaneous', 'drinks quality', 'drinks style_options', 'restaurant general' and 'food style_options'.
- The 'sentiment polarity' refers to the degree of positivity, negativity or neutrality expressed in the opinion towards a particular aspect or feature of a product or service, and the available polarities inlcudes: 'positive', 'negative' and 'neutral'.

Recognize all sentiment elements with their corresponding aspect terms, aspect categories, opinion terms and sentiment polarity in the following text with the format of [('aspect term', 'opinion term', 'aspect category', 'sentiment polarity'), ...]: 

Text: never again !
Sentiment Elements: [('null', 'never', 'restaurant general', 'bad')]

Text: the food was mediocre at best but it was the horrible service that made me vow never to go back .
Sentiment Elements: [('food', 'mediocre', 'food quality', 'bad'), ('service', 'horrible', 'service general', 'bad')]

Text: we had the lobster sandwich and it was fantastic .
Sentiment Elements: [('lobster sandwich', 'fantastic', 'food quality', 'great')]

Text: they have it all -- great price , food , and service .
Sentiment Elements: [('null', 'great', 'restaurant prices', 'great'), ('food', 'great', 'food quality', 'great'), ('service', 'great', 'service general', 'great')]

Text: they even scoop it out nice ( for those on a diet ) not too much not to little .
Sentiment Elements: [('null', 'nice', 'food style_options', 'great')]

Text: also it 's great to have dinner in a very romantic and comfortable place , the service it 's just perfect ... they 're so frendly that we never want to live the place !
Sentiment Elements: [('place', 'romantic', 'ambience general', 'great'), ('place', 'comfortable', 'ambience general', 'great'), ('service', 'perfect', 'service general', 'great')]

Text: my friend from milan and myself were pleasantly surprised when we arrived and everyone spoke italian .
Sentiment Elements: [('null', 'pleasantly surprised', 'restaurant miscellaneous', 'great')]

Text: i had their eggs benedict for brunch , which were the worst in my entire life , i tried removing the hollondaise sauce completely that was how failed it was .
Sentiment Elements: [('eggs benedict', 'worst', 'food quality', 'bad')]

Text: the food is authentic italian - delicious !
Sentiment Elements: [('food', 'authentic italian', 'food quality', 'great'), ('food', 'delicious', 'food quality', 'great')]

Text: a little pricey but it really hits the spot on a sunday morning !
Sentiment Elements: [('null', 'pricey', 'restaurant prices', 'bad'), ('null', 'hits the spot', 'restaurant general', 'great')]
\end{lstlisting}
\end{figure*}

\section{Additional Comparison with Single-view Prompting}
\label{sec:appendix:svp}

Previously, we conducted ablation studies with a single view on several representative tasks and datasets in both full-data and low-resource settings (see \S \ref{sec:results:ablation}). Figure \ref{fig:num-view} illustrates that incorporating multiple views leads to significant improvements, especially in low-resource settings. Additionally, Table \ref{tab:ablation} showcases the performance degradation resulting from the use of two single-view aggregation strategies.

Here, we present additional comparisons with single-view prompting (designated as SvP) on all tasks, i.e., selecting the best single view for training and inference. We experiment using three single-view selection strategies:
1) \textbf{random};
2) \textbf{heuristic}: In our pre-experiments, we find that elements ranked ahead of the top selected orders are mostly free-form terms '$[\mathtt{A}]$' and '$[\mathtt{O}]$', which have higher uncertainty than '$[\mathtt{C}]$' and '$[\mathtt{S}]$'. Therefore, we propose using the "$[\mathtt{A}][\mathtt{O}][\mathtt{C}][\mathtt{S}]$" order heuristically;
3) \textbf{rank}: choosing a view for each dataset based on the score described in \S \ref{subsubsec:method:order-select}.

As depicted in Table \ref{tab:svp-result}, the results show that the SvP methods, whether employing a ranking strategy or a heuristic order chosen through pre-experimentation, exhibit limited improvement over a random order (+0.37 and +0.67). By permuting and aggregating results from multiple views, MvP significantly outperforms SvP across all tasks (+1.90). These consistent improvements further elucidate the efficacy of MvP.

\begin{table*}[ht]
\centering
\begin{tabular}{c|lr|ccc}
\toprule
\multicolumn{1}{c|}{Task} &
  Dataset &
  \#Cat &
  \begin{tabular}[c]{@{}c@{}}Train \\ (POS/NEU/NEG)\end{tabular} &
  \begin{tabular}[c]{@{}c@{}}Dev\\ (POS/NEU/NEG)\end{tabular} &
  \begin{tabular}[c]{@{}c@{}}Test\\ (POS/NEU/NEG)\end{tabular} \\ \midrule
\multicolumn{1}{c|}{\multirow{3}{*}{ASQP}} &
  Rest15 &
  13 &
  \begin{tabular}[c]{@{}c@{}}834\\ 1,005/34/315\end{tabular} &
  \begin{tabular}[c]{@{}c@{}}209\\ 252/14/81\end{tabular} &
  \begin{tabular}[c]{@{}c@{}}537\\ 453/37/305\end{tabular} \\ \cmidrule{2-6} 
\multicolumn{1}{c|}{} &
  Rest16 &
  13 &
  \begin{tabular}[c]{@{}c@{}}1,264\\ 1,369/62/558\end{tabular} &
  \begin{tabular}[c]{@{}c@{}}316\\ 341/23/143\end{tabular} &
  \begin{tabular}[c]{@{}c@{}}544\\ 584/40/177\end{tabular} \\ \midrule
\multirow{3}{*}{ACOS} &
  Laptop &
  121 &
  \begin{tabular}[c]{@{}c@{}}2,934\\ 2,583/227/1,364\end{tabular} &
  \begin{tabular}[c]{@{}c@{}}326\\ 279/24/137\end{tabular} &
  \begin{tabular}[c]{@{}c@{}}816\\ 716/65/380\end{tabular} \\ \cmidrule{2-6} 
 &
  Restaurant &
  13 &
  \begin{tabular}[c]{@{}c@{}}1,530\\ 1,656/95/733\end{tabular} &
  \begin{tabular}[c]{@{}c@{}}171\\ 180/12/69\end{tabular} &
  \begin{tabular}[c]{@{}c@{}}583\\ 668/44/205\end{tabular} \\ \midrule
\multirow{8}{*}{ASTE} &
  Laptop14 &
  - &
  \begin{tabular}[c]{@{}c@{}}906\\ 817/126/517\end{tabular} &
  \begin{tabular}[c]{@{}c@{}}219\\ 169/36/141\end{tabular} &
  \begin{tabular}[c]{@{}c@{}}328\\ 364/63/116\end{tabular} \\ \cmidrule{2-6} 
 &
  Rest14 &
  - &
  \begin{tabular}[c]{@{}c@{}}1,266\\ 1,692/166/480\end{tabular} &
  \begin{tabular}[c]{@{}c@{}}310\\ 404/54/119\end{tabular} &
  \begin{tabular}[c]{@{}c@{}}492\\ 773/66/155\end{tabular} \\ \cmidrule{2-6} 
 &
  Rest15 &
  - &
  \begin{tabular}[c]{@{}c@{}}605\\ 783/25/205\end{tabular} &
  \begin{tabular}[c]{@{}c@{}}148\\ 185/11/53\end{tabular} &
  \begin{tabular}[c]{@{}c@{}}322\\ 317/25/143\end{tabular} \\ \cmidrule{2-6} 
 &
  Rest16 &
  - &
  \begin{tabular}[c]{@{}c@{}}857\\ 1,015/50/329\end{tabular} &
  \begin{tabular}[c]{@{}c@{}}210\\ 252/11/76\end{tabular} &
  \begin{tabular}[c]{@{}c@{}}326\\ 407/29/78\end{tabular} \\ \midrule
\multirow{3}{*}{TASD} &
  Rest15 &
  13 &
  \begin{tabular}[c]{@{}c@{}}1,120\\ 1,198/53/403\end{tabular} &
  \begin{tabular}[c]{@{}c@{}}10\\ 6/0/7\end{tabular} &
  \begin{tabular}[c]{@{}c@{}}582\\ 454/45/346\end{tabular} \\ \cmidrule{2-6} 
 &
  Rest16 &
  13 &
  \begin{tabular}[c]{@{}c@{}}1,708\\ 1,657/101/749\end{tabular} &
  \begin{tabular}[c]{@{}c@{}}29\\ 23/1/20\end{tabular} &
  \begin{tabular}[c]{@{}c@{}}587\\ 611/44/204\end{tabular} \\ 
  \bottomrule
\end{tabular}
\caption{Dataset statistics for various tasks. \#Cat refers to the number of aspect categories in the set. POS, NEU, and NEG denote the number of positive, neutral and negative quads or triplets respectively.}
\label{tab:data-sta-all}
\end{table*}
\end{document}